\documentclass[letterpaper]{article} 
\usepackage{preprint}   
\nocopyright            
\usepackage{times}  
\usepackage{helvet}  
\usepackage{courier}  
\usepackage[hyphens]{url}  
\usepackage{graphicx} 
\usepackage[numbers]{natbib}  
\usepackage{caption} 
\usepackage{algorithm}
\usepackage{algorithmic}

\usepackage{newfloat}
\usepackage{listings}
\DeclareCaptionStyle{ruled}{labelfont=normalfont,labelsep=colon,strut=off} 
\floatstyle{ruled}
\newfloat{listing}{tb}{lst}{}
\floatname{listing}{Listing}
\title{SAGA: Scene-Aware, Goal-Evolving Agents\\
       for Long-Horizon Strategy Game Planning}

\author{
    Tianyu Jin\textsuperscript{\rm 1},
    Shuo Chen\textsuperscript{\rm 2,*},
    Yida Wang\textsuperscript{\rm 1},
    Liuyu Xiang\textsuperscript{\rm 1,*},
    Yingzhuo Liu\textsuperscript{\rm 1},
    Yexin Li\textsuperscript{\rm 2},
    Peipei Li\textsuperscript{\rm 1},
    Zhaofeng He\textsuperscript{\rm 1}
}
\affiliations{
    \textsuperscript{\rm 1}Beijing University of Posts and Telecommunications \quad
    \textsuperscript{\rm 2}Beijing Institute for General Artificial Intelligence (BIGAI)\\
    \textsuperscript{*}Corresponding authors.\quad
    \texttt{chenshuo@bigai.ai}, \texttt{xiangliuyu@bupt.edu.cn}
}

\usepackage[utf8]{inputenc}
\usepackage[T1]{fontenc}
\usepackage{booktabs}
\usepackage{amsfonts}
\usepackage{amssymb}
\usepackage{amsmath}
\usepackage{nicefrac}
\usepackage{microtype}
\usepackage{xcolor}
\usepackage{multirow}
\usepackage{graphicx}
\usepackage{multirow}
\usepackage{makecell}   
\usepackage{array}
\usepackage{colortbl}
\usepackage{pifont}     
\usepackage{dblfloatfix} 
\usepackage{afterpage}   
\usepackage{subcaption}  
\usepackage[most]{tcolorbox}  
\usepackage{tabularx}         
\usepackage{placeins}         

\newcommand{\cmark}{\textcolor{green!60!black}{\ding{51}}}  
\newcommand{\xmark}{\textcolor{red}{\ding{55}}}             

\newcommand{\methodname}{SAGA}
\makeatletter
\@ifundefined{ack}{}{}
\makeatother

\begin{document}

\maketitle

\begin{abstract}
{
Grand-strategy games such as Civilization pose a distinctive
long-horizon planning problem: an agent must divide one shared
resource pool among six competing domains---technology, government,
diplomacy, city development, expansion, and military---under partial
observability, with no feedback except a delayed final score.
Current LLM agents fall short in three ways:
1)~they cannot infer spatial relations from raw coordinates;
2)~they allocate resources poorly, because feeding the entire growing
state into one prompt and planning all domains in a single output
diffuses attention and biases decisions toward urgent events; and
3)~they cannot improve, as the delayed score gives no signal within or
across games.
We present \methodname{}, an LLM multi-agent framework pairing one
mechanism with each weakness:
(i)~a \textbf{Map-Semantic Scene Graph} turning coordinates into
per-entity statements of distance, direction, and threat;
(ii)~a \textbf{Tool-Augmented Planner} that retrieves only the state a
decision needs---cutting the order of magnitude of its input---and
issues a separate plan per domain to six specialist
\textbf{controllers}, so urgent events do not derail long-term plans;
and
(iii)~a \textbf{Dual-Horizon Feedback Loop} setting short-term goals
during play and distilling each game into lessons for the next.
On CivRealm, a Civilization-style benchmark, \methodname{} leads five
LLM baselines on mean final score and is the only method significantly
ahead of all of them on city development---the first investment
baselines sacrifice---with 27\% fewer output tokens; with cross-game
learning it scores highest after five games, and its fifth game
consistently surpasses its first across four maps.
Our code is available at
\url{https://github.com/Kazecloudk/SAGA-Scene-Aware-Goal-Evolving-Agents-for-Long-Horizon-Strategy-Game-Planning}.
}
\end{abstract}


\section{Introduction}
\label{sec:intro}

\begin{figure}[t]
  \centering
  \includegraphics[width=0.84\columnwidth]{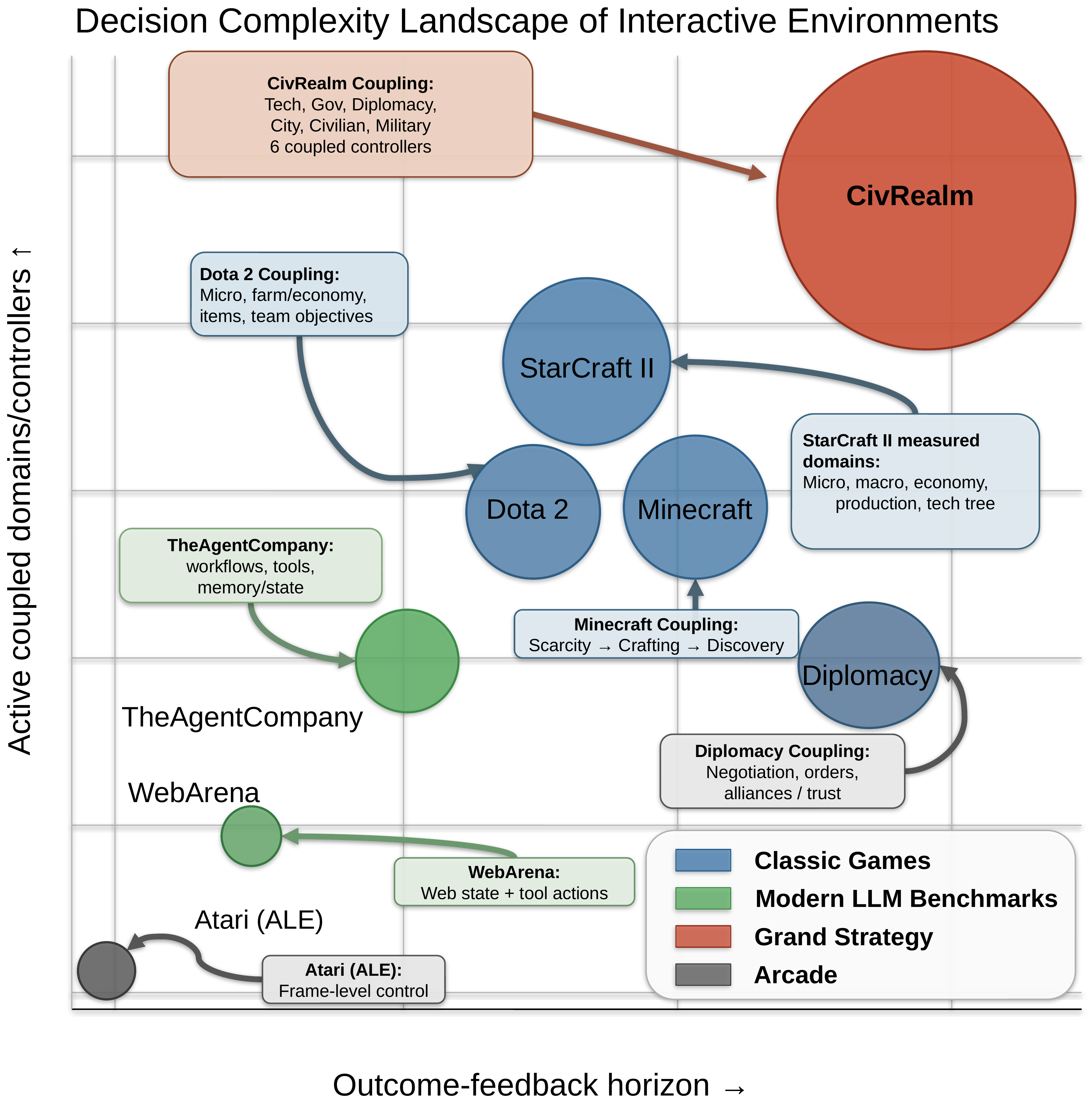}
  \caption{Decision-complexity landscape. Rightward, outcome feedback
    arrives over ever longer horizons (Challenge~3); upward, more
    decision domains are coupled (Challenge~2); the map-scale
    observations behind both axes give rise to Challenge~1. CivRealm
    occupies the extreme regime, and \methodname{} pairs one mechanism
    with each challenge (\S\ref{sec:method}).}
  \label{fig:freeciv_env}
\end{figure}

Long-horizon planning in Grand-strategy games such as
Civilization---where an agent must concurrently manage multi-domain
decision-making including technology development,
government/economic/diplomacy policy, city production, civilian
expansion, and military operations across hundreds of turns under
imperfect information and lagged score feedback---represents a
fundamental frontier for LLM-based planning research.
We study this problem with CivRealm~\cite{civrealm2024}, a
decision-making benchmark built on FreeCiv, an open-source counterpart
of the Civilization series.
CivRealm combines six tightly coupled decision domains (technology,
government, diplomacy, city management, civilian units, and military
units), outcome feedback that arrives only as a lagged aggregate score
over hundred-turn games, hard irreversible mechanical rules, and a state
space that keeps expanding as the game unfolds
(Figure~\ref{fig:freeciv_env}).

Three challenges make this setting difficult for LLM agents;
Figure~\ref{fig:freeciv_env} situates them along the environment's two
complexity axes.
\textbf{Challenge~1: spatial understanding.}
The environment reports every unit and city as raw tile coordinates. To
act, a language model must recover from these numbers what a human
player reads off the map at a glance---which enemy threatens which
city, from which direction, and how far away---by arithmetic over
coordinate pairs, a step at which current LLMs are unreliable. Agents
therefore behave as if blind to the map, missing approaching threats
and expanding in arbitrary directions.
\textbf{Challenge~2: resource allocation under a shared budget.}
The core difficulty of grand strategy is not any single domain but
their common purse: all six domains draw on one shared pool of city
productivity---itself a product of city count, population, food
reserves, terrain, and the infrastructure built on it---so every
investment in one domain is a cut to the others.
Existing agents handle this allocation problem poorly, for two
compounding reasons.
On the input side, they feed the entire, ever-growing game state into
every prompt---beyond 15k tokens per step in the late game---so the
signals relevant to a given decision are diluted among a mass of
irrelevant state, a regime in which LLM attention is known to
degrade~\cite{cos2024,epicstar2025}.
On the output side, they plan every domain in one single generation,
which systematically over-reacts to urgent, salient events (an
approaching army) while starving slow-payoff investments such as
granaries and marketplaces~\cite{hima2025,optimus2}---the very city
development that compounds into long-run capability.
\textbf{Challenge~3: long-horizon credit assignment.}
The sole objective signal is the final score of a hundreds-of-turns game.
Within a game, agents receive no intermediate indication of whether a
strategy is working; across games, prior methods retrieve and replay past
trajectories~\cite{epicstar2025,evolver2025} without attributing outcomes
to the decisions that caused them, so the same strategic mistakes recur
from episode to episode.

We introduce \textbf{\methodname}, an LLM multi-agent framework that pairs
one mechanism with each challenge, instantiated and validated on CivRealm.
The \textbf{Map-Semantic Scene Graph} restores spatial understanding
(Challenge~1) by precomputing exactly the facts that coordinate
arithmetic would otherwise have to recover: each turn it builds a graph
over units and cities and converts each entity's neighbourhood into a
few plain sentences of distance, direction, and threat.
This spatial context is new information that no baseline provides, and
we add it in a controlled way: each controller receives only its own
entity's neighbourhood, never a description of the whole graph, so the
added text stays proportional to local surroundings rather than to the
size of the map.
The \textbf{Tool-Augmented Planner} attacks the allocation problem
(Challenge~2) on both of its sides: on the input side it starts from a
compact strategic digest and retrieves fine-grained state only when a
decision requires it, cutting by an order of magnitude the volume of
state the planner must reason over; on the output side it issues a
separate plan for each domain, routed to one of six dedicated
\emph{specialist controllers}---so a sudden emergency reshapes the
military plan without derailing long-term plans already in motion.
The \textbf{Dual-Horizon Feedback Loop} supplies the missing feedback
(Challenge~3) at two timescales: within a game, periodic goal generation
provides measurable intermediate goals between score updates; across
games, a structured post-mortem attributes each outcome to the decisions
behind it and folds the lessons into a strategic prior for the next
game, without manual reward engineering.

\noindent\textbf{Contributions.}\;
\textbf{(C1)}~We propose \methodname{}, whose three mechanisms---the
Map-Semantic Scene Graph, the Tool-Augmented Planner with decoupled
specialist controllers, and the Dual-Horizon Feedback Loop---respectively
target spatial understanding, resource allocation under a shared budget,
and long-horizon credit assignment.
\textbf{(C2)}~Against five competitive LLM baselines on CivRealm,
\methodname{} attains the highest mean on five of six development metrics,
with its statistically strongest advantage on infrastructure, while cutting
output tokens by 27\% relative to the strongest baseline
(Table~\ref{tab:main}; significance tests in Appendix~\ref{app:stats}).
\textbf{(C3)}~With the same cross-game evolution module attached to every
method, \methodname{} reaches the highest final score after five
successive games (Figure~\ref{fig:evolution}), and across four
independently generated maps its fifth game consistently surpasses its
first in score and in military, city, and infrastructure development
(Figure~\ref{fig:radar}); ablations confirm each mechanism contributes
independently (Table~\ref{tab:ablation}).

\section{Related Work}
\label{sec:related}

\paragraph{Complex Strategic Environments.}
Decision-making benchmarks span cooperative~\cite{hanabi2020,meltingpot2021},
competitive~\cite{dota2berner2019,alphastar2019}, partially
observable~\cite{minedojo2022,diplomacy2019}, and
task-oriented~\cite{webarena2024,osworld2024,gamabench2025,theagentcompany2025}
settings, each stressing a different capability.
Real-time strategy games such as StarCraft~II stress fast reactive control
over a compact production loop~\cite{alphastar2019}; open-world games such
as Minecraft chain long tasks with weak coupling between concurrent
objectives~\cite{minedojo2022}.
Grand strategy occupies a distinct regime: many mutually constraining
decision domains, irreversible mechanics, and outcomes that materialize
only after hundreds of turns.
No existing benchmark simultaneously satisfies all ten criteria this regime
demands (Appendix~\ref{app:env_comparison}), motivating
CivRealm~\cite{civrealm2024}, whose position on both complexity axes
Figure~\ref{fig:freeciv_env} quantifies, as our evaluation environment.

\paragraph{Large-Language-Model-based Agents for Decision Making.}
ReAct~\cite{react2023} established the foundational paradigm of interleaving
chain-of-thought reasoning with tool calls.
Subsequent work extended this to open-world curricula~\cite{voyager2023},
multi-agent software pipelines~\cite{metagpt2024}, distributed
coordination~\cite{collab2025}, and embodied perception
bottlenecks~\cite{embeddedbench2025}.
Planning architectures for long-horizon strategy games include
CoS~\cite{cos2024}, which compresses interaction history into rolling text
summaries; Optimus-2~\cite{optimus2}, which conditions action selection on
phase milestones; and HIMA~\cite{hima2025}, which fuses proposals from
parallel advisor agents into a unified plan.
None of these planners represents spatial relations between game entities
explicitly: CoS and HIMA reason over aggregate textual summaries
with no geometric structure, and Mastaba~\cite{civrealm2024}, the native
CivRealm agent, condenses the map by pooling $15{\times}15$-tile regions into
fixed blocks---a resolution reduction that preserves density statistics but
discards entity-level relations such as which unit threatens which city.
\methodname{} instead encodes typed entity-to-entity relations whose
semantics match the decisions to be made (\S\ref{sec:scene});
Section~\ref{sec:experiments} details every baseline as re-implemented on our
shared infrastructure.

\paragraph{LLM Agents with Reflection and Self-Evolution Capabilities.}
Reflexion~\cite{reflexion2023} and Self-Refine~\cite{selfrefine2023} introduced
verbal self-criticism for iterative output refinement within a single task.
Across tasks, EvolveR~\cite{evolver2025} distills interaction trajectories
into reusable strategic principles---the spirit closest to our cross-game
outer loop---and EpicStar~\cite{epicstar2025} retrieves episodic memories of
similar past states; related self-improvement pipelines optimize agent
trajectories or multi-agent reasoning chains~\cite{seagent2025,sirius2025}.
These methods transfer \emph{what happened}, but none attributes game
outcomes to the strategic decisions that caused them, so retrieved experience
remains heuristic replay.
\methodname{} instead derives a causal post-mortem from every completed
game and maintains an explicitly revised strategic prior, with within-game
goals that make the attribution measurable (\S\ref{sec:feedback}).

\section{The \methodname{} Framework}
\label{sec:method}

\subsection{System Overview}

\begin{figure*}[t]
  \centering
  \includegraphics[width=0.78\textwidth]{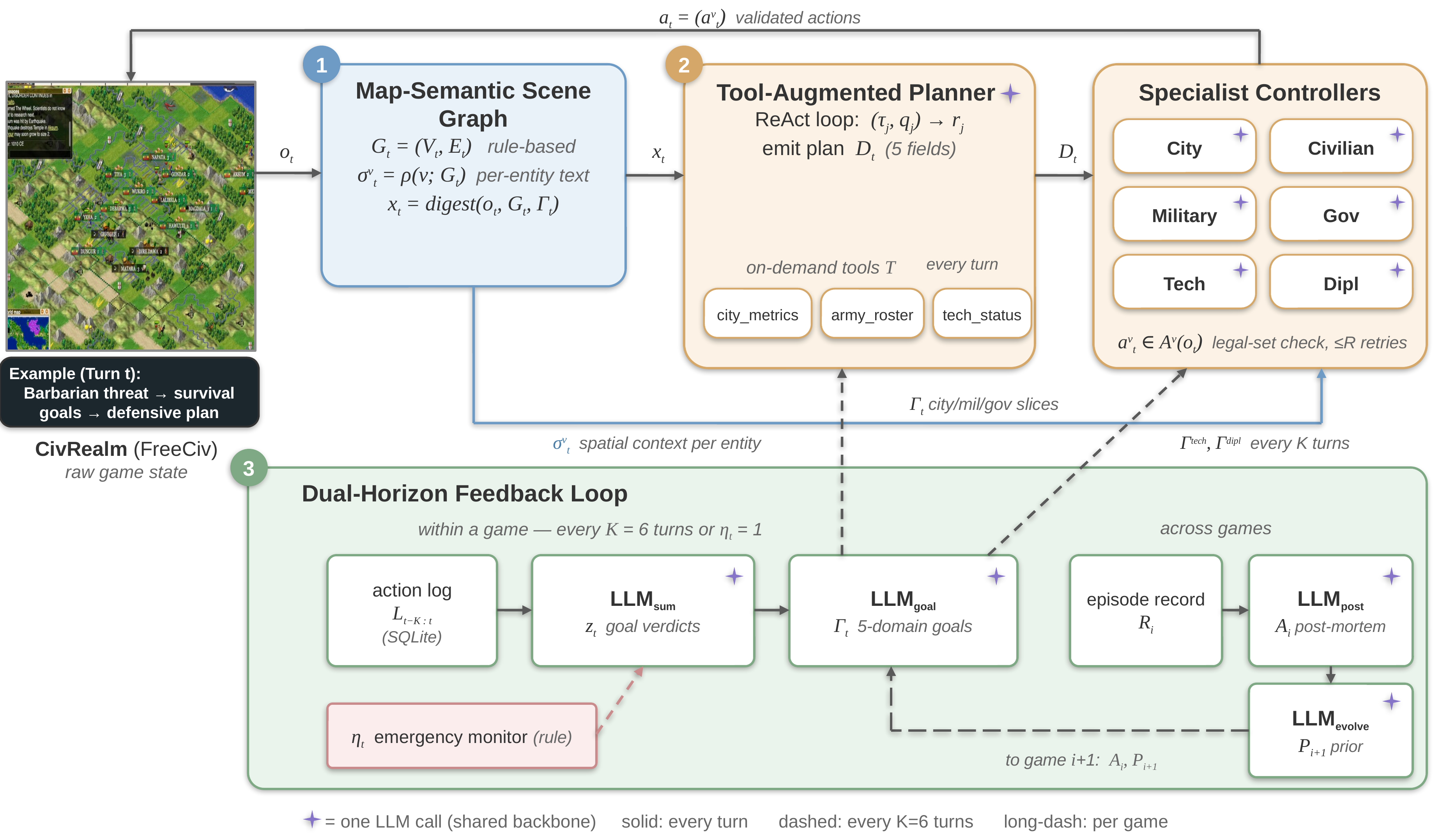}
  \caption{\textbf{\methodname{} system architecture.}
    Each turn, deterministic rules compile the observation $o_t$ into
    the scene graph $\mathcal{G}_t$, per-entity renderings
    $\sigma^{v}_t$, and a bounded digest $x_t$ (\S\ref{sec:scene}).
    The Tool-Augmented Planner queries on-demand tools inside a ReAct
    loop and emits the structured plan $\mathcal{D}_t$, which is routed
    together with the cycle-level goals $\Gamma_t$ to six specialist
    controllers whose actions are validated against the legal sets
    $\mathcal{A}^{v}(o_t)$ (\S\ref{sec:planner}).
    The Dual-Horizon Feedback Loop refreshes $z_t$ and $\Gamma_t$ every
    $K{=}6$ turns or upon an emergency $\eta_t$, and distills each
    finished game into a post-mortem $A_i$ and a strategic prior
    $P_{i+1}$ (\S\ref{sec:feedback}).
    Star badges mark LLM calls
    (Eqs.~\ref{eq:react}, \ref{eq:controller},
    \ref{eq:sum}--\ref{eq:evolve}); every other component is a
    deterministic rule.}
  \label{fig:architecture}
\end{figure*}

Addressing the three challenges of \S\ref{sec:intro}, \methodname{} employs
three tightly integrated mechanisms (Figure~\ref{fig:architecture}):
the \textbf{Map-Semantic Scene Graph}~(\S\ref{sec:scene}) targets spatial
understanding (Challenge~1);
the \textbf{Tool-Augmented Planner} with decoupled specialist
controllers~(\S\ref{sec:planner}) targets resource allocation under the
shared budget, including the context growth that aggravates it
(Challenge~2); and
the \textbf{Dual-Horizon Feedback Loop}~(\S\ref{sec:feedback}) targets
long-horizon credit assignment (Challenge~3).
The environment is formalized as a POMDP with a factored six-domain
action space in Appendix~\ref{sec:prelim}.
Throughout, $t$ indexes turns within a game, $i$ indexes successive
games, $o_t$ denotes the observation at turn $t$, and $\mathcal{G}_t$
the scene graph built from it.
Every \methodname{} agent is the same frozen backbone LLM specialized
only by a role prompt and a typed output schema:
$y = \mathrm{LLM}_{r}(\pi_r;\,\cdot)$ denotes one backbone call under
role $r$, whose output is decoded into $r$'s schema by constrained JSON
parsing.
Here $\pi_r$ is an \emph{expert-designed role prompt}, and it is a
first-class input rather than boilerplate: it encodes the human
strategic priors that tell the role what to attend to, how to compare
one game against another, and what makes a sub-goal well-formed.
Without such priors an LLM cannot reliably perform the strategic
abstraction, goal setting, or cross-game reflection these roles
require; the complete prompt for every role is given in
Appendix~\ref{app:prompts}.
No weights are updated anywhere in the framework; every other map below
(graph construction, rendering, digesting, routing, validation) is a
deterministic rule.
For full pseudocode, see Appendix~\ref{app:algorithm}.

\subsection{Map-Semantic Scene Graph}
\label{sec:scene}

\paragraph{Construction.}
At every turn $t$, \methodname{} builds
$\mathcal{G}_t = (\mathcal{V}_t, \mathcal{E}_t)$ from the observation $o_t$
by deterministic rules---no learned parameters and no manual annotation;
rebuilding from scratch each turn keeps the graph consistent with the
visible state at negligible cost ($<$10\,ms on CPU;
Appendix~\ref{app:impl}).
The nodes
$\mathcal{V}_t = \mathcal{C}^{a}_t \cup \mathcal{U}^{a}_t \cup
\mathcal{C}^{e}_t \cup \mathcal{U}^{e}_t$
are the agent's own (\emph{allied}) cities and units and all visible
\emph{enemy} cities and units; node $v$ carries its entity type and position
$p(v) \in \mathbb{Z}^2$.
$d(v, v')$ denotes Manhattan distance, and
$c^{*}(v) = \arg\min_{c} d(v, c)$ the nearest city of $v$'s own side.

\paragraph{Relation types.}
Edges come in four kinds, each answering a different question a player asks about the map.
The edge set
$\mathcal{E}_t = \mathcal{E}^{\mathrm{anc}}_t \cup
\mathcal{E}^{\mathrm{prox}}_t \cup \mathcal{E}^{\mathrm{atk}}_t \cup
\mathcal{E}^{\mathrm{thr}}_t$
comprises four relation types, each consumed by a specific class of decision.
\emph{Anchor edges} $\{(u, c^{*}(u)) : u \in \mathcal{U}^{a}_t\}$ link every
allied unit to its nearest allied city with no distance cutoff, preserving
navigation context in unexplored territory (movement, settlement).
\emph{Proximity edges}
$\{(v, v') : v, v' \in \mathcal{C}^{a}_t \cup \mathcal{U}^{a}_t,\,
d(v, v') \le r_{p}\}$
connect allied entities within radius $r_{p}$ (local coordination such as
garrisoning).
\emph{Attack edges}
$\{(u, c) : u \in \mathcal{U}^{a,\mathrm{mil}}_t,\, c \in \mathcal{C}^{e}_t\}$
link military units to visible enemy cities (offensive targeting).
\emph{Threat edges}
$\{(e, v^{*}(e)) : e \in \mathcal{U}^{e}_t,\,
d(e, c^{*}(e)) > r_{\mathrm{inf}}\}$
fire when an enemy unit moves beyond radius $r_{\mathrm{inf}}$ of its own
nearest city, a pattern consistent with an advancing attack, and alert the
nearest allied entity $v^{*}(e)$ (defensive response).

\paragraph{From graph to text.}
Each edge $e = (v, v')$ carries the label
$\ell(e) = \bigl(\beta(d(v, v')),\, \theta(v, v')\bigr)$, where $\beta$
discretizes distance into \emph{close} ($<$3), \emph{medium} ($<$10), and
\emph{far} ($\ge$10) and $\theta$ is the eight-way compass direction of
$p(v') - p(v)$; band thresholds and radii $(r_{p}, r_{\mathrm{inf}})$ match
the game's unit vision and movement ranges, refined in pilot runs and fixed
across all experiments (Appendix~\ref{app:impl}).
A text generator $\rho$ turns the edges touching a single entity $v$
into a few short sentences,
$\sigma^{v}_{t} = \rho(v;\mathcal{G}_t)$, delivered only to that
entity's controller in Eq.~\ref{eq:controller} (examples in
Appendix~\ref{app:prompts}).
This spatial description is information no baseline provides, so it
necessarily adds tokens; what matters is that the addition stays
controlled. Each controller is given only the sentences for its own
entity---never a description of the whole graph---so the text added to
any single decision is proportional to that entity's immediate
surroundings rather than to the map or the empire as a whole.

\subsection{Tool-Augmented Planning and Decoupled Execution}
\label{sec:planner}

\paragraph{On-demand state retrieval.}
Baseline planners receive the full serialized state before every decision,
an input that grows linearly with owned cities, units, and explored tiles
and exceeds 15k tokens per step in late game, diluting the signals that
matter for any single decision (\S\ref{sec:intro}).
\methodname{} instead starts each turn from a compact digest
\begin{equation}
  x_t \;=\; \mathrm{digest}\bigl(o_t,\, \mathcal{G}_t,\, \Gamma_t\bigr)
      \;=\; \bigl(m_t,\; g_t,\; u_t,\; \alpha_t\bigr),
  \label{eq:digest}
\end{equation}
where $m_t$ aggregates per-domain empire metrics (counts, rates, and one
status line per city), $g_t$ reports progress toward the active
intermediate goals $\Gamma_t$ (\S\ref{sec:feedback}), $u_t$ is a
directional expansion summary derived from $\mathcal{G}_t$, and
$\alpha_t$ lists active threat alerts.
Every component is an aggregate over domains rather than a listing of
entities, so $|x_t|$ grows far more slowly than the raw state it
replaces. We do not claim a hard bound---an empire may in principle
keep founding cities---but the digest lowers by roughly an order of
magnitude the volume of text the planner must reason over before it
decides what to look at next.

\paragraph{ReAct planning loop.}
The central planner is a ReAct-style agent~\cite{react2023} equipped
with the on-demand query tools
$\mathcal{T} = \{\texttt{city\_metrics}, \texttt{army\_roster},
\texttt{tech\_status}\}$
(Table~\ref{tab:tools}, Appendix~\ref{app:impl}), each a deterministic
read-only view of the current state.
Every turn it interleaves reasoning with tool calls: for
$j = 1, 2, \ldots$,
\begin{equation}
  \begin{aligned}
    (\tau_j, q_j) &= \mathrm{LLM}_{\mathrm{plan}}\bigl(\pi_{\mathrm{plan}};\, x_t,\, \bar\sigma_t,\, \Gamma_t,\, z_t,\,
        (q_k, r_k)_{k<j}\bigr),\\
    r_j &= \tau_j(q_j;\, o_t),
  \end{aligned}
  \label{eq:react}
\end{equation}
where $\tau_j \in \mathcal{T}$ is the selected tool, $q_j$ its
arguments, $r_j$ the returned state slice, $\bar\sigma_t$ the
threat-relevant subset of the scene renderings of \S\ref{sec:scene},
and $z_t$, $\Gamma_t$ the current cycle summary and goals of
\S\ref{sec:feedback} (the planner receives the goal slices of the
fast domains it steers in Eq.~\ref{eq:routing}).
The loop terminates at the iteration whose output is not a further
query but the structured plan
\begin{equation}
  \mathcal{D}_t \;=\; \bigl(
      \mathcal{D}^{\mathrm{city}}_t,\;
      \mathcal{D}^{\mathrm{gar}}_t,\;
      \mathcal{D}^{\mathrm{civ}}_t,\;
      \mathcal{D}^{\mathrm{mil}}_t,\;
      \mathcal{D}^{\mathrm{gov}}_t\bigr),
  \label{eq:plan}
\end{equation}
five schema-disjoint fields: per-city production orders, per-city
garrison assignments, per-unit directives for civilian and for field
military units, and a government/tax directive.
Fine-grained state thus enters the context only when a decision
requires it: each decision is grounded in the exact slice it needs,
and the input scales with what the planner asks for rather than with
everything the empire happens to contain.

\paragraph{Two-timescale directive routing.}
Directives reach the six specialist domains of
Eq.~\ref{eq:action_space} from two sources on different timescales:
the per-turn plan $\mathcal{D}_t$ steers the fast-evolving domains,
while for the slowly-evolving technology and diplomacy domains the
cycle-level goals $\Gamma_t$, refreshed every $K$ turns
(\S\ref{sec:feedback}), serve directly as the standing directive:
\begin{equation}
  \delta^{d}_t \;=\;
  \begin{cases}
    \mathcal{D}^{d}_t,
      & d \in \{\text{city},\, \text{civ}\},\\[2pt]
    \mathcal{D}^{\mathrm{gar}}_t \oplus \mathcal{D}^{\mathrm{mil}}_t,
      & d = \text{mil},\\[2pt]
    \mathcal{D}^{\mathrm{gov}}_t \oplus \Gamma^{\mathrm{gov}}_t,
      & d = \text{gov},\\[2pt]
    \Gamma^{d}_t,
      & d \in \{\text{tech},\, \text{dipl}\},
  \end{cases}
  \label{eq:routing}
\end{equation}
with $\oplus$ denoting prompt concatenation.
This partition directly prevents urgent events from starving long-term
investment: an urgent military situation reshapes
$\mathcal{D}^{\mathrm{mil}}_t$ without overwriting
$\mathcal{D}^{\mathrm{city}}_t$ or unsettling the standing research
target $\Gamma^{\mathrm{tech}}_t$. In a unified output, by contrast,
the same emergency reasoning repeatedly overrides production planning,
which is the mechanism behind the systematic over-militarization we
observe across all LLM baselines (\S\ref{sec:ablation}).

\paragraph{Decoupled controller execution.}
Each active entity $v$ (city or unit) and each domain-level function
(technology, government, diplomacy) is handled by the specialist
controller of its domain $d(v)$, which maps its directive to one action
from the entity's currently legal action set $\mathcal{A}^{v}(o_t)$:
\begin{equation}
  a^{v}_t \;=\; \mathrm{LLM}_{d(v)}\bigl(\pi_{d(v)};\,
      \delta^{d(v)}_t,\; \sigma^{v}_t,\; \omega^{v}_t,\; h^{v}_t
    \bigr) \;\in\; \mathcal{A}^{v}(o_t),
  \label{eq:controller}
\end{equation}
where $\omega^{v}_t$ is the entity's own observation slice (position,
strength, production state, and its enumerated legal actions),
$\sigma^{v}_t$ its scene rendering from \S\ref{sec:scene} (spatial
grounding for civilian tile actions and military tactics; empty for the
domain-level controllers), and $h^{v}_t$ its recent action history.
Controllers run in parallel across entities; the diplomacy controller,
which mutates shared negotiation state, is serialized
(Appendix~\ref{app:impl}).
As an implementation safeguard---not a contribution of this
work---actions outside $\mathcal{A}^{v}(o_t)$ trigger a re-prompt (up
to $R{=}5$ attempts before a no-op), and city production is restricted
to the technology-unlocked subset.

\subsection{Dual-Horizon Feedback Loop}
\label{sec:feedback}

\paragraph{Short-term inner loop (within a game).}
Let $\mathcal{L}_{t-K:t}$ denote the persisted log of digests and
per-controller decisions over the last $K$ turns.
At every cycle boundary---every $K{=}6$ turns, a cadence balancing
planning overhead against responsiveness, or immediately upon an
emergency ($\eta_t{=}1$, below)---a summarization agent audits the
elapsed cycle against the goals $\Gamma_{t^-}$ set at the previous
boundary,
\begin{equation}
  z_t \;=\; \mathrm{LLM}_{\mathrm{sum}}\bigl(\pi_{\mathrm{sum}};\,
      \Gamma_{t^-},\; \mathcal{L}_{t-K:t},\; x_t\bigr),
  \label{eq:sum}
\end{equation}
returning one verdict per domain goal (\emph{success} /
\emph{failure} / \emph{in-progress}, each tied to the actions
responsible) plus corrective notes---affirming directions on track
(e.g., early settler production) and flagging missteps (e.g., military
spending crowding out infrastructure).
A goal generator then refreshes the goals, conditioned on this verdict
and on the cross-game artifacts $A_{i-1}, P_i$ of the outer loop below:
\begin{equation}
  \Gamma_t \;=\; \mathrm{LLM}_{\mathrm{goal}}\bigl(\pi_{\mathrm{goal}};\,
      x_t,\; z_t,\; A_{i-1},\; P_i,\; \eta_t\bigr),
  \label{eq:goal}
\end{equation}
emitting a strategic posture (expansion, consolidation, offense, or
survival) and one time-bounded objective per strategic domain,
$\Gamma_t = (\Gamma^{\mathrm{city}}_t, \Gamma^{\mathrm{mil}}_t,
\Gamma^{\mathrm{tech}}_t, \Gamma^{\mathrm{dipl}}_t,
\Gamma^{\mathrm{gov}}_t)$,
e.g.,~\emph{produce three settlers and two granaries within ten turns;
research Alphabet within seven}.
Between boundaries, $z_t$ and $\Gamma_t$ persist unchanged; progress
toward $\Gamma_t$ is folded into the digest component $g_t$ of
Eq.~\ref{eq:digest} on every subsequent turn, converting the interval
between score updates into measurable feedback.

\paragraph{Emergency replanning.}
A rule-based monitor sets $\eta_t{=}1$ when hostile military units are
newly sighted near allied cities (within four tiles), when a controlled
city falls to opponents, or on first contact with a new opponent, with
a five-turn refractory period to prevent thrashing.
An emergency forces an immediate cycle boundary: Eqs.~\ref{eq:sum}
and~\ref{eq:goal} re-run with the alert appended, and the planner
receives it through $\alpha_t$, overriding the periodic schedule.
Without this interrupt, the planner keeps issuing expansion directives
while the empire is under active attack.

\begin{table*}[t]
  \caption{\textbf{Main quantitative results} on Map Seed 2025 (base planner
    only, identical initial states): mean $\pm$ std at turn~150 over $n{=}10$
    games per method; $\uparrow$ = higher is better. Best mean per metric in
    \textbf{bold}; $^{\dagger}$ marks the only metric on which
    \methodname{}'s lead is significant against \emph{all} five baselines
    (two-sided Mann--Whitney $p<0.05$, Holm-corrected); the Score lead is
    significant over three of five baselines and within noise against CoS
    and HIMA. Per-comparison tests, medians, and win rates appear in
    Appendix~\ref{app:stats}.}
  \label{tab:main}
  \centering
  \footnotesize
  \setlength{\tabcolsep}{3pt}
  \begin{tabular}{lcccccc}
    \toprule
    \textbf{Method} &
      \textbf{Score}\,$\uparrow$ &
      \textbf{Tech}\,$\uparrow$ &
      \textbf{Military}\,$\uparrow$ &
      \textbf{Cities}\,$\uparrow$ &
      \textbf{Buildings}\,$\uparrow$ &
      \textbf{Diplomacy}\,$\uparrow$ \\
    \midrule
    CoS~\cite{cos2024}
      & 47.8 $\pm$ 20.3 & 13.3 $\pm$ 3.4
      & 24.4 $\pm$ 36.9 & 5.5 $\pm$ 6.9 & 7.5 $\pm$ 5.6 & \textbf{41.5 $\pm$ 22.0} \\
    EpicStar~\cite{epicstar2025}
      & 35.4 $\pm$ 21.5 & 10.2 $\pm$ 3.2
      & 23.1 $\pm$ 28.9 & 5.1 $\pm$ 8.4 & 4.1 $\pm$ 5.2 & 38.5 $\pm$ 17.8 \\
    Optimus-2~\cite{optimus2}
      & 41.5 $\pm$ 10.1 & 12.6 $\pm$ 1.9
      & 35.9 $\pm$ 39.2 & 4.1 $\pm$ 3.4 & 1.9 $\pm$ 1.8 & 16.5 $\pm$ 13.6 \\
    HIMA~\cite{hima2025}
      & 43.7 $\pm$ 22.3 & 10.8 $\pm$ 3.5
      & 28.9 $\pm$ 26.9 & 7.5 $\pm$ 6.5 & 3.2 $\pm$ 2.3 & 19.5 $\pm$ 17.7 \\
    Mastaba~\cite{civrealm2024}
      & 39.0 $\pm$ 14.2 & 12.2 $\pm$ 3.5
      & 7.8 $\pm$ 9.4 & 4.0 $\pm$ 6.8 & 2.3 $\pm$ 2.6 & 7.0 $\pm$ 9.5 \\
    \midrule
    \textbf{\methodname{} (Ours)}
      & \textbf{60.7 $\pm$ 14.0} & \textbf{13.4 $\pm$ 2.5}
      & \textbf{59.3 $\pm$ 40.8} & \textbf{8.9 $\pm$ 4.5} & \textbf{16.3 $\pm$ 10.5}$^{\dagger}$ & 39.5 $\pm$ 18.0 \\
    \bottomrule
  \end{tabular}
\end{table*}

\paragraph{Long-term outer loop (across games).}
At the end of game $i$, a post-game analyst distills the complete
episode record $\mathcal{R}_i$---the digest trajectory subsampled every
ten turns, per-domain action and production histories, the inner
loop's cycle verdicts, threat and city-loss events, and final
metrics---into a structured post-mortem
\begin{equation}
  A_i \;=\; \mathrm{LLM}_{\mathrm{post}}\bigl(\pi_{\mathrm{post}};\,\mathcal{R}_i\bigr),
  \label{eq:post}
\end{equation}
which attributes decisive outcomes in successful and failed games
alike to the decisions behind them
(e.g.,~\emph{``early government transition enabled sustained economic
dominance''}; \emph{``technology disadvantage caused the military
collapse at turn~107''}) and closes with per-domain recommendations
for the next game.
Before game $i{+}1$, a pre-game strategist rebuilds the persistent
strategic prior from the most recent post-mortems,
\begin{equation}
  P_{i+1} \;=\; \mathrm{LLM}_{\mathrm{evolve}}\bigl(\pi_{\mathrm{evolve}};\,
      A_{i-w+1}, \ldots, A_i\bigr), \qquad w{=}4,
  \label{eq:evolve}
\end{equation}
tracing which strategy revisions helped, which failed, and which
failures persist; the sliding window keeps $|P_{i+1}|$ bounded over
long evolution runs.
$P_{i+1}$ conditions goal generation (Eq.~\ref{eq:goal}) throughout
game $i{+}1$, so per-game analyses accumulate into a transferable prior
without manual reward engineering.

\section{Experiments}
\label{sec:experiments}

\subsection{Experimental Setup}

\paragraph{Environment \& Implementation.}
All experiments use the CivRealm environment~\cite{civrealm2024} with
FreeCiv 2.6 as backend.
Each episode runs at most 150 turns on a 23$\times$23 Classical map against
two built-in AI opponents plus active barbarians and pirates, under a fixed
global map seed (2025).
To ensure a strictly fair comparison, all baselines and our method use the
same \textbf{Doubao-Seed-1.8} backbone through the same API endpoint.
Each method is evaluated over \textbf{10} independent runs, and we report
the mean and standard deviation; hyperparameter justifications are in
Appendix~\ref{app:hyperparameters}.

\paragraph{Baselines.}
We compare against five LLM-based planning baselines:
CoS~\cite{cos2024} (compresses turn history into rolling summaries for context management),
EpicStar~\cite{epicstar2025} (retrieves similar past episodes as cross-game strategic memory),
Optimus-2~\cite{optimus2} (conditions action selection on expand/consolidate phase milestones),
HIMA~\cite{hima2025} (aggregates multi-advisor strategic council into a unified plan), and
Mastaba~\cite{civrealm2024} (condenses the map into a compact rule-based state sketch).
All baselines share the same observation-to-text wrapper, ReAct-style
controller infrastructure, and LLM backbone;
only the planner architecture and strategy module differ.
A qualitative paradigm comparison across five design dimensions is provided
in Table~\ref{tab:comparison} (Appendix~\ref{app:comparison}).

\paragraph{Baseline Adaptation Protocol.}
Most baselines target other environments and do not run natively on
CivRealm, so we re-implement all of them---including Mastaba, the
native CivRealm agent---on one shared infrastructure (observation
adapter, tool suite, six controllers with action-mask validation,
logging, and, when enabled, the evolution chain); only the central
planner differs, so performance gaps are attributable to the planner
architecture alone (details in Appendix~\ref{app:impl}).

\paragraph{Evaluation Metrics.}
We assess seven dimensions---Score, Technology, Military, Cities,
Buildings, Diplomacy, and Token Consumption (input/output over 150
turns)---all read at turn~150; definitions and computation details are
in Appendix~\ref{app:metrics}.

\subsection{Main Results}

\paragraph{Overall performance.}
Table~\ref{tab:main} reports all six methods on Map Seed 2025 (base planner
only, no cross-game evolution module; $n{=}10$ per method). \methodname{}
attains the highest mean on five of the six metrics, with statistical
strength that varies across metrics (full tests in
Appendix~\ref{app:stats}).
The clearest advantage is on \textbf{infrastructure}: \methodname{} builds
far more than any baseline (Buildings $16.3$ vs.\ $\leq7.5$), and this is
the \emph{only} metric on which it significantly beats \emph{all} five
baselines (two-sided Mann--Whitney $p<0.05$, Holm-corrected), with a
head-to-head win rate of $78$--$98\%$---the fraction of same-seed game
pairs in which \methodname{}'s final value exceeds the baseline's
(Appendix~\ref{app:stats}).
On \textbf{Score}, \methodname{} posts the highest mean ($60.7\pm14.0$)
\emph{and} the lowest variance among the capable methods (std $14.0$ vs.\
CoS $20.3$, HIMA $22.3$); the lead is significant over the weaker three
baselines but, given CivRealm's heavy-tailed scores, within noise against
the two strongest---CoS ($47.8$, $p{=}0.06$) and HIMA ($43.7$,
$p{=}0.10$)---over which \methodname{} nonetheless wins $\geq72\%$ of
head-to-head games.
Cities and Military lead in the mean but reach significance against only a
subset of baselines, so we do not rest our claims on them.
Technology is a statistical tie ($13.4$ vs.\ CoS $13.3$): \methodname{}
matches the field's breadth while selecting high-value prerequisites on
demand, which converts an equal technology count into markedly more
Buildings and a higher Score.
On token cost (Table~\ref{tab:token}), \methodname{} trades higher
input ($258$k) for far lower output ($109$k) than the two strongest
baselines, CoS ($\sim$149k) and HIMA ($\sim$219k); output tokens
dominate price and decoding latency, and the reduction reflects
compact, well-formed plans rather than verbose, hedged ones (full
analysis in Appendix~\ref{sec:app_token}).

\begin{figure}[t]
  \centering
  \includegraphics[width=\columnwidth]{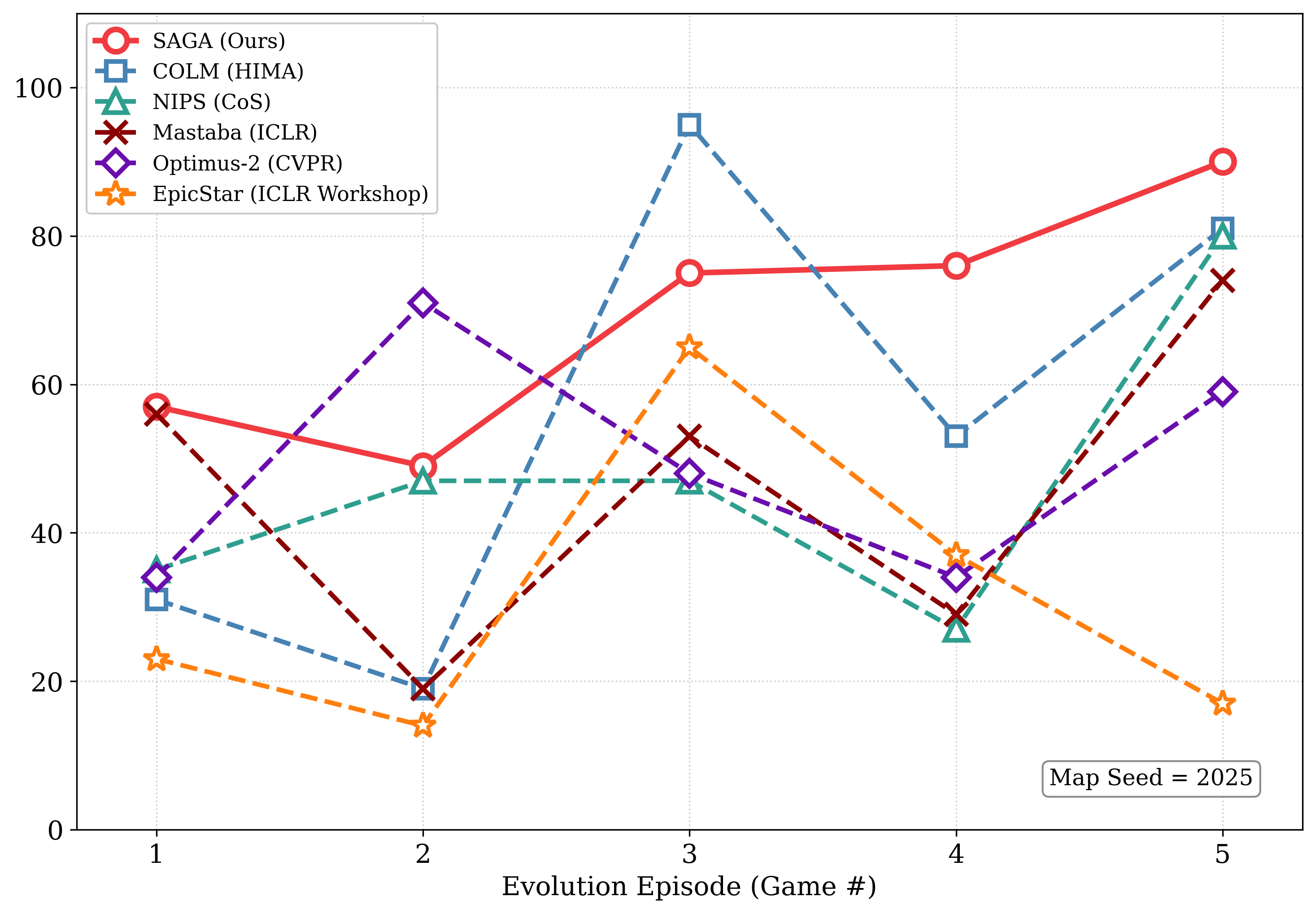}
  \caption{\textbf{Cross-game evolution.} All methods carry the same
    cross-game evolution module over five successive games (Map Seed 2025).
    Baseline trajectories oscillate between episodes; \methodname{} leads
    throughout and reaches the highest final score ($90$). Figure~\ref{fig:radar} extends this protocol
    across map seeds.}
  \label{fig:evolution}
\end{figure}

\paragraph{Architecture-agnostic scalability of the evolution module.}
Figure~\ref{fig:evolution} isolates the evolution module: attached to
the baselines under identical conditions, it lifts nearly all of them
substantially from Game~1 to Game~5 (HIMA $31\to81$; CoS $35\to80$;
\methodname{} $57\to90$), acting as an architecture-agnostic
enhancement that distills transferable strategic knowledge.
The exception is EpicStar ($23\to17$), whose TF-IDF episodic retrieval
fails to surface relevant memories in CivRealm's high-dimensional
state space: evolution amplifies a capable backbone but cannot
compensate for poor retrieval precision.

\begin{table*}[t]
  \caption{\textbf{Ablation study} of intra-game components: each variant
    removes one component from the full system ($n{=}10$ Game-1 runs, Map
    Seed 2025, mean~$\pm$~std). Cross-game evolution modules are validated
    separately (Figure~\ref{fig:radar}).}
  \label{tab:ablation}
  \centering
  \footnotesize
  \setlength{\tabcolsep}{6pt}
  \resizebox{\textwidth}{!}{%
  \begin{tabular}{lcccccccc}
    \toprule
    \textbf{Variant} &
      \textbf{Score}\,$\uparrow$ &
      \textbf{Tech}\,$\uparrow$ &
      \textbf{Mil.}\,$\uparrow$ &
      \textbf{Cities}\,$\uparrow$ &
      \textbf{Bldg.}\,$\uparrow$ &
      \textbf{Dipl.}\,$\uparrow$ &
      \textbf{In\,(k)}\,$\downarrow$ &
      \textbf{Out\,(k)}\,$\downarrow$ \\
    \midrule
    Full \methodname{}                             & 60.7 $\pm$ 14.0 & 13.4 $\pm$ 2.5 & 59.3 $\pm$ 40.8 & 8.9 $\pm$ 4.5 & 16.3 $\pm$ 10.5 & 39.5 $\pm$ 18.0 & $258{\pm}54$ & $109{\pm}14$ \\
    \quad w/o Scene Graph                          & 45.3 $\pm$ 9.8 & 12.3 $\pm$ 1.6 & 38.3 $\pm$ 23.2 & 5.0 $\pm$ 2.1 & 10.3 $\pm$ 5.7 & 45.5 $\pm$ 27.2 & $301{\pm}85$ & $94{\pm}22$ \\
    \quad w/o Tools                                & 30.9 $\pm$ 9.5 & 9.4 $\pm$ 2.1 & 30.0 $\pm$ 20.5 & 2.6 $\pm$ 1.3 & 3.8 $\pm$ 4.2 & 24.0 $\pm$ 25.1 & $133{\pm}10$ & $88{\pm}24$ \\
    \quad w/o Inner Loop                           & 45.1 $\pm$ 8.6 & 5.9 $\pm$ 2.2 & 67.5 $\pm$ 28.0 & 9.4 $\pm$ 2.4 & 5.9 $\pm$ 3.9 & 42.5 $\pm$ 26.9 & $206{\pm}52$ & $78{\pm}13$ \\
    \bottomrule
  \end{tabular}%
  }
\end{table*}

\begin{figure}[t]
  \centering
  \setlength{\tabcolsep}{1pt}
  \begin{tabular}{cc}
    \includegraphics[width=0.49\columnwidth]{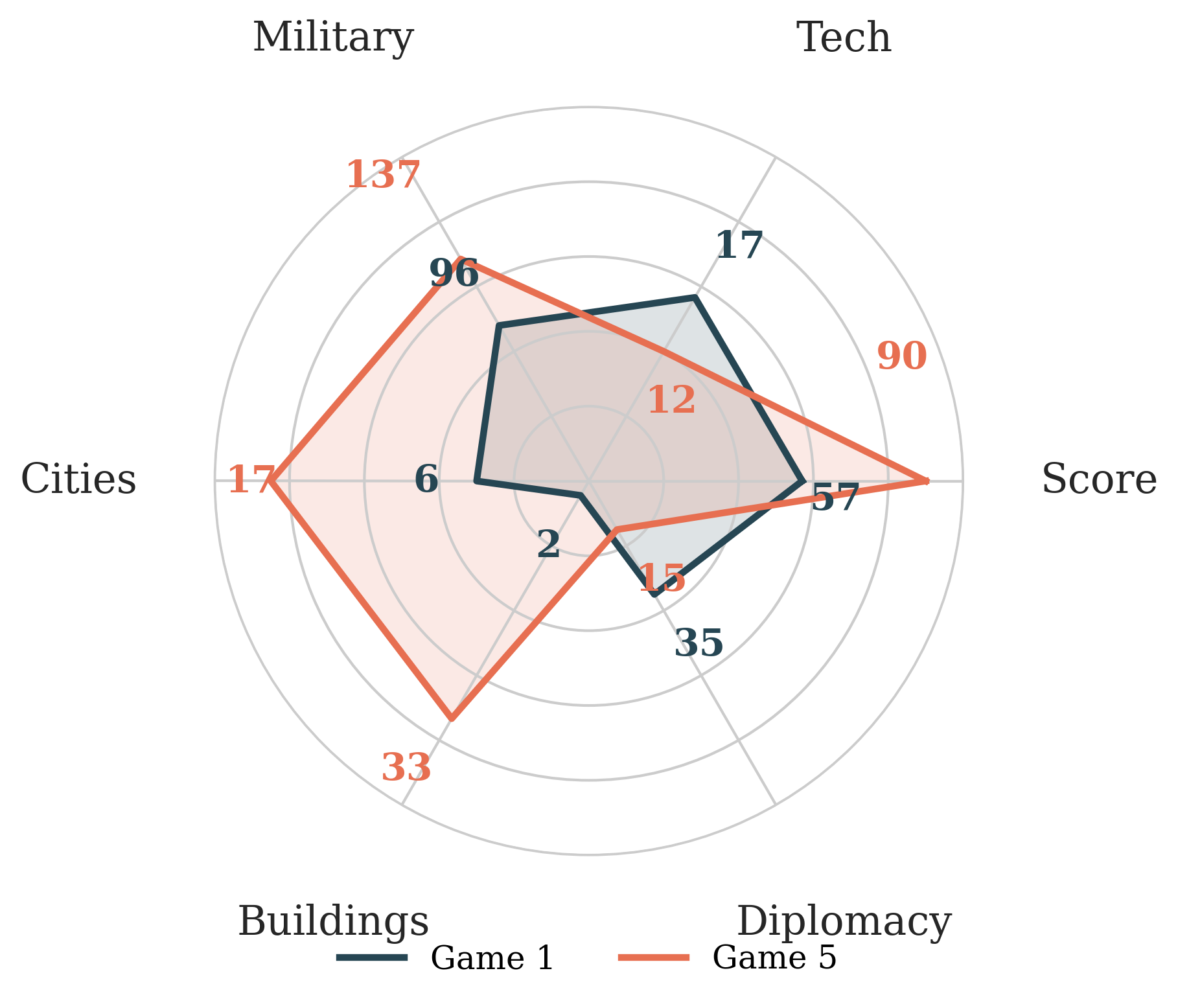} &
    \includegraphics[width=0.49\columnwidth]{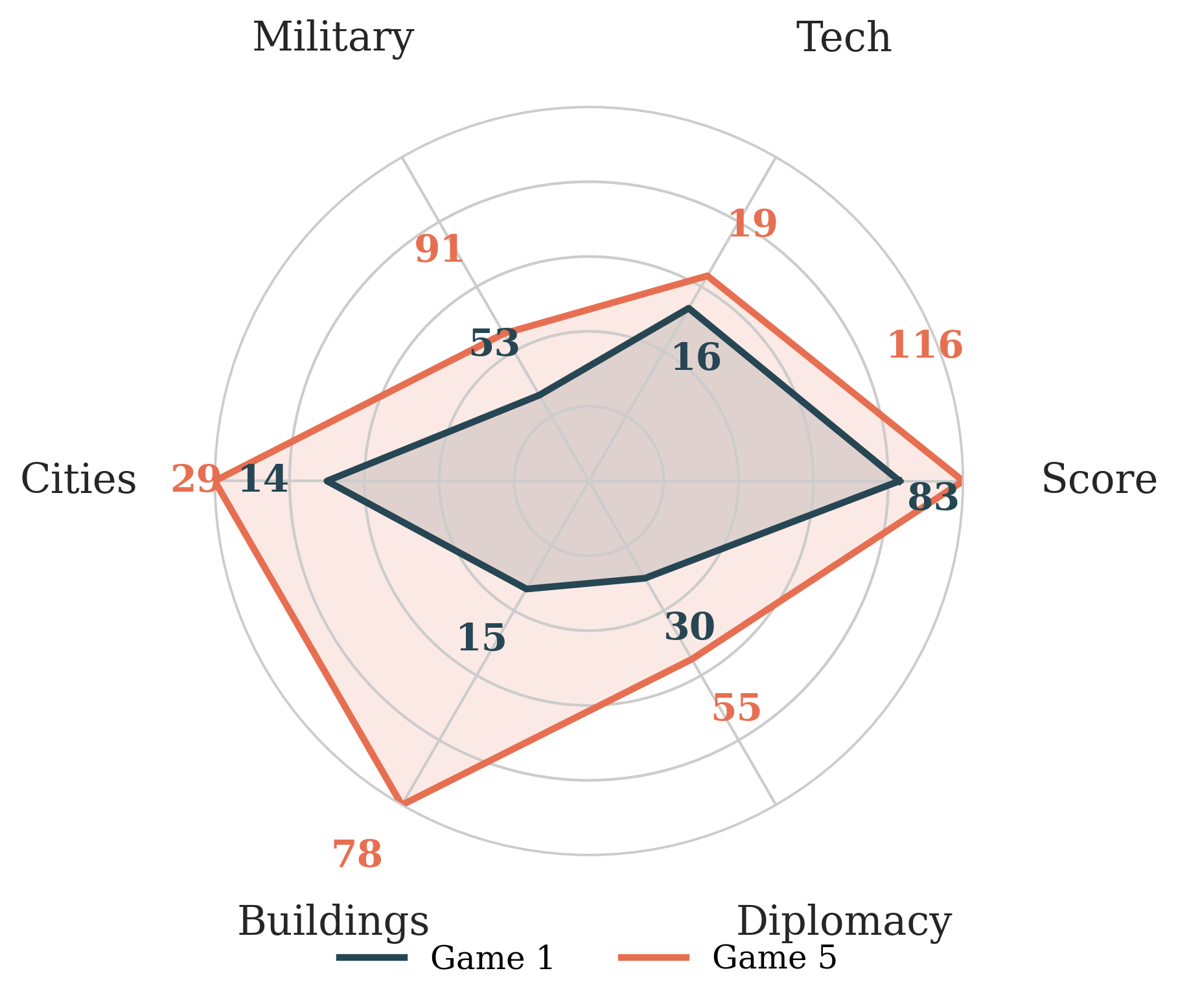} \\
    \small (a) Seed 2025 & \small (b) Seed 2026 \\[-0.3ex]
    \includegraphics[width=0.49\columnwidth]{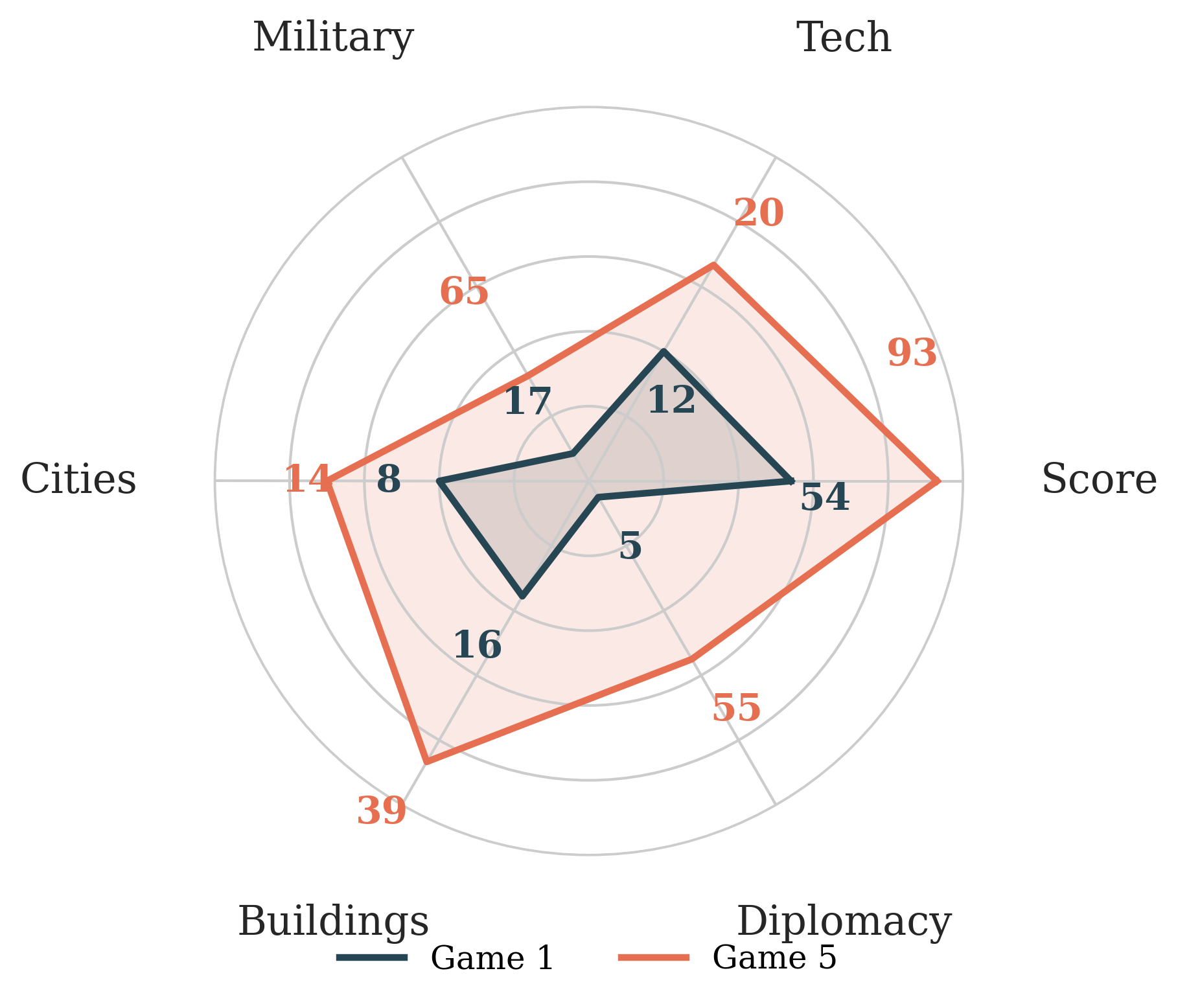} &
    \includegraphics[width=0.49\columnwidth]{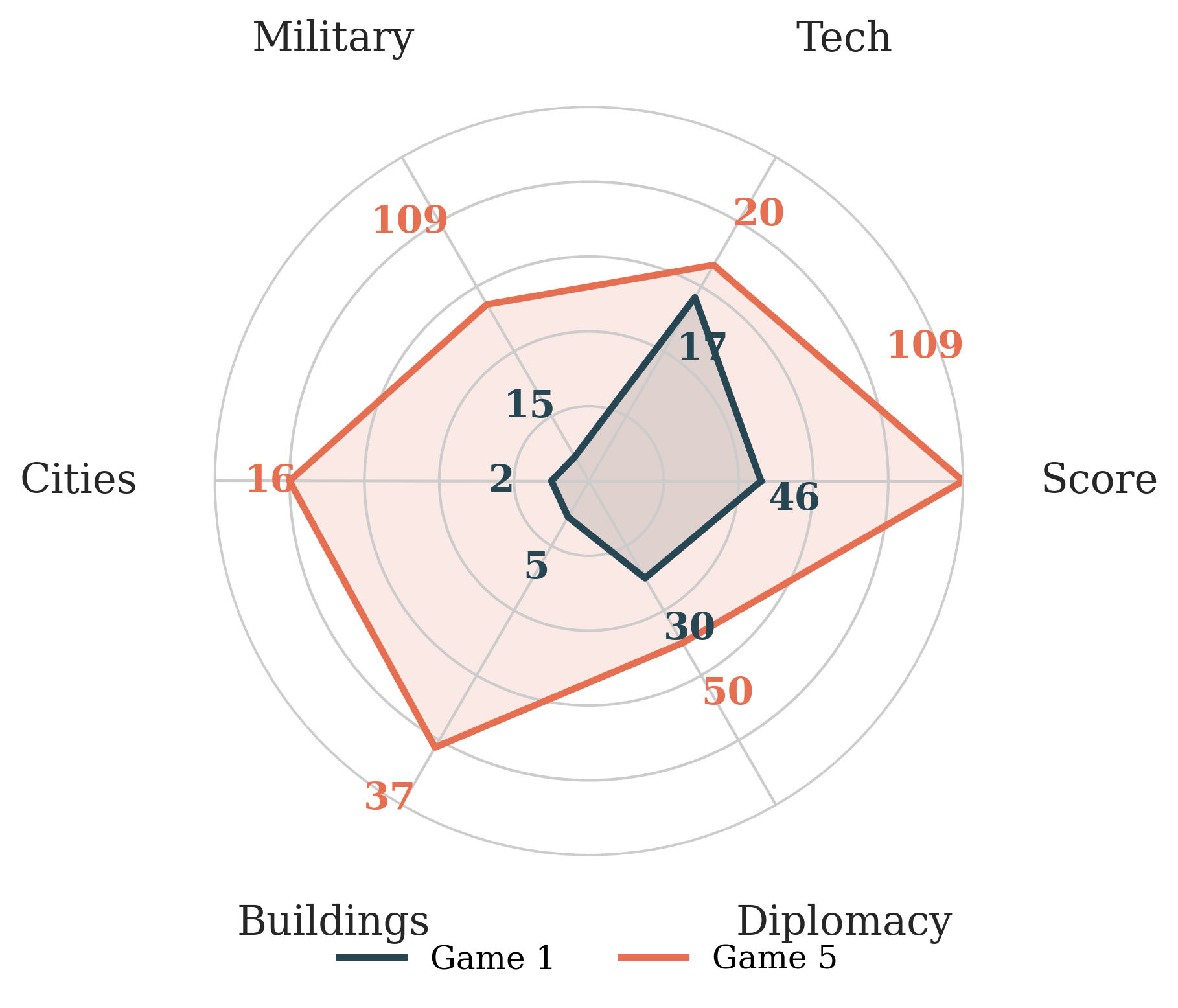} \\
    \small (c) Seed 2027 & \small (d) Seed 2028 \\
  \end{tabular}
  \caption{\textbf{Multi-seed evolution validation.} Radar charts across
    four map seeds compare \methodname{}'s Game~1 (dark) with its Game~5
    (red) under the protocol of Figure~\ref{fig:evolution}. In every seed
    the polygon expands on Score, Military, Cities, and Buildings,
    indicating that the accumulated strategic prior transfers to
    structurally different maps.}
  \label{fig:radar}
\end{figure}

\paragraph{Stable knowledge exploitation requires a capable backbone.}
All baseline trajectories in Figure~\ref{fig:evolution} oscillate
sharply between games; \methodname{} faces the same stochasticity yet
leads from the start and finishes highest.
We attribute this to a knowledge-utilization bottleneck: injected
priors help only if the decision layer can ground directives in the
scene, acquire precise state on demand, and translate long-horizon
goals into verifiable sub-goals---exactly the capabilities the
baselines lack, which is why they oscillate or collapse while
\methodname{} converts the same priors into a consistent lead.

\paragraph{Cross-seed generalization.}
Across four independent map seeds (Figure~\ref{fig:radar}), the Game-5
radar polygon is substantially larger than its Game-1 counterpart in
every seed, driven by Score, Military, Cities, and Buildings: the
accumulated strategic principles transfer to structurally different
maps.
This validates the cross-game modules at the system level, so
Table~\ref{tab:ablation} ablates only intra-game components.

\subsection{Ablation Studies}
\label{sec:ablation}

To isolate \methodname{}'s architectural contributions, we ablate its
intra-game framework along its three components: the Map-Semantic Scene
Graph, the on-demand tools, and the inner-loop feedback
(Table~\ref{tab:ablation}).

\textbf{Impact of disabling the Scene Graph.}
Replacing the scene graph with a raw global observation dump degrades
every development metric (Score $-25\%$, Cities $-44\%$, Buildings
$-37\%$, Military $-35\%$) while input tokens \emph{rise} from $258$k
to $301$k: the planner ingests more raw state yet decides worse.

\textbf{Impact of disabling on-demand tools.}
Removing tool calling forces the planner to parse dense, unqueried
text and is the most severe degradation of all variants (Score
$-49\%$, Tech $-30\%$, Buildings $-77\%$): precise on-demand state is
essential to resource tracking and strategic execution.

\textbf{Effect of removing the inner-loop feedback.}
Dismantling the $K$-turn cycle drops Score by $26\%$ and, most
diagnostically, collapses Tech ($-56\%$) and Buildings ($-64\%$):
without time-bounded goals the planner drifts between tasks,
abandoning research chains mid-way and switching production before
buildings finish.
The residually high Military ($67.5$) shows reactive decisions still
fire; what is lost is sustained multi-turn commitment.

\textbf{Token cost is justified; the scene graph even saves it.}
Output tokens fall in every ablation ($109\text{k}\!\to\!94/88/78$k),
but each drop costs performance, so the extra output buys useful
decisions rather than verbosity.
Removing tools or the inner loop lowers input ($133$k/$206$k)---their
context is the justified price of precise planning---whereas removing
the scene graph \emph{raises} input to $301$k: the graph is an input
compressor, not an overhead.

\paragraph{Backbone generalization.}
To test whether these results depend on a strong backbone, we repeat the
single-game protocol with the weaker GPT-4o-mini in place of Doubao-Seed-1.8
against the two strongest baselines (CoS, HIMA), holding all other
infrastructure fixed. \methodname{} leads on all six metrics, with $78\%$
higher Score than the stronger baseline ($51.2$ vs.\ $28.8$), roughly
$2.4\times$ the Tech, and over $3\times$ the Buildings of either baseline,
so the capability ordering of Table~\ref{tab:main} survives a substantial
downgrade of the reasoning engine; full results and analysis are given in
Appendix~\ref{sec:backbone}.

\subsection{Qualitative Analysis and Discussion}
\label{sec:analysis}

The cross-game evolution loop improves several dimensions at once: the
post-game analyst extracts structured causal chains diagnosing
underperformance across expansion, development, and defense, and the
injected priors produce the broad polygon expansion of
Figure~\ref{fig:radar}.
Baselines given the same evolution chain oscillate or stagnate,
confirming that structured intra-game execution is a prerequisite for
exploiting cross-game knowledge.

\paragraph{Limitations and future work.}
Three limitations point to concrete next steps.
First, the hierarchy duplicates context across the goal-setting,
planning, and summarization agents; shared memory would cut this
overhead, as would distilling the repetitive entity controllers into
lightweight reinforcement-learning policies and reserving the backbone
for strategic decisions.
Second, a fixed seed does not remove CivRealm's randomness---barbarian
spawns drive the large deviations of every method (Military std
$>$40)---which ten runs mitigate but cannot eliminate.
Third, 150 turns are too short to balance every trade-off, so
single-game overcorrection, most visibly over-militarization, is hard
to avoid; the outer loop targets exactly this, within the limits of
backbone capability (Appendix~\ref{app:failure_cases}).
Full-length games would sharpen the allocation problem and test
whether cross-game priors keep transferring, and a vision-language
model reading the map directly could replace our handcrafted spatial
text.

\section{Conclusion}
\label{sec:conclusion}

We presented \methodname, a scene-aware, tool-augmented LLM multi-agent
framework for long-horizon strategy planning.
By replacing full state dumps with on-demand queries, grounding unit
decisions in per-entity spatial context, and coupling within-game goal
setting with cross-game evolution, \methodname{} attains the highest
mean on five of six metrics, a statistically robust lead on city
development, and $\geq72\%$ head-to-head Score wins against the two
strongest baselines.
The results suggest that the bottleneck in LLM strategic planning lies
less in raw reasoning capability than in \emph{information
architecture}: how observations are structured, decisions decoupled,
and feedback accumulated.



\twocolumn[%
  \begin{center}
    \vspace*{0.3em}
    {\LARGE\bfseries Technical Appendix}\par
    \vspace{1.4em}
  \end{center}
]
\appendix

\section{Preliminaries}
\label{sec:prelim}

\paragraph{Partially Observable Markov Decision Process.}
We model the FreeCiv strategy game as a discrete-time Partially
Observable Markov Decision Process (POMDP), defined by the tuple
$\mathcal{M} = \langle \mathcal{S}, \mathcal{O}, \mathcal{A}, T, R, \gamma \rangle$.
$\mathcal{S}$ is the global game state (map tiles, units, cities,
technology, government, diplomacy); $\mathcal{O}$ is the observation space,
a strict subset of $\mathcal{S}$ under persistent fog-of-war.
The joint action space decomposes over six controller-aligned strategic domains:
\begin{equation}
  \mathcal{A} = \mathcal{A}^{tech} \times \mathcal{A}^{gov} \times
  \mathcal{A}^{dipl} \times \mathcal{A}^{city} \times
  \mathcal{A}^{civilian} \times \mathcal{A}^{military}.
  \label{eq:action_space}
\end{equation}
$T\colon \mathcal{S} \times \mathcal{A} \to \Delta(\mathcal{S})$ is the
stochastic transition kernel; $R\colon \mathcal{S} \times \mathcal{A} \to \mathbb{R}$
is the lagged aggregate reward (the in-game civilization score); and
$\gamma \in (0,1)$ is the discount factor.
The agent policy $\pi\colon \mathcal{O}^{*} \to \mathcal{A}$ targets high
cumulative reward $\mathbb{E}\bigl[\sum_{t=0}^{T}\gamma^{t}R(s_t, a_t)\bigr]$.
We adopt this POMDP purely as a \emph{problem formulation} that motivates
\methodname{}'s architecture; we do not learn or optimize $\pi$ by
reinforcement learning and make no optimality claim.

\paragraph{FreeCiv as a POMDP Instance.}
FreeCiv~\cite{civrealm2024} instantiates $\mathcal{M}$ at large combinatorial
scale: 87 technology types, 68 building types, 52 unit types, 6 government
forms, and 5 diplomatic states; the joint state space grows from roughly
$10^{15}$ configurations in the early game to $10^{650}$ in advanced stages,
and the action space from $10^{4}$ to $10^{166}$~\cite{civrealm2024}.
Observations $o_t \in \mathcal{O}$ arrive across six modalities (terrain
grid, units, cities, technology, government, diplomacy) and are translated
into structured natural language by the benchmark's observation-to-text
interface.
This combinatorial scale, combined with the strict information asymmetry
between $\mathcal{S}$ and $\mathcal{O}_t$ across hundreds of turns on
stochastically generated maps, motivates the architectural choices of
Section~\ref{sec:method}.

\section{Notation}
\label{app:notation}

Table~\ref{tab:notation} summarizes the symbols used in
Section~\ref{sec:method} and Algorithm~\ref{alg:saga}.

\begin{table}[t]
\caption{\textbf{Notation.} Symbols of the \methodname{} framework,
  with the defining equation or section.}
\label{tab:notation}
\centering
\small
\setlength{\tabcolsep}{4pt}
\begin{tabular}{@{}l p{0.60\columnwidth}@{}}
\toprule
\textbf{Symbol} & \textbf{Meaning} \\
\midrule
$t$;\ $i$ & turn index within a game; game index \\
$o_t$ & observation at turn $t$ \\
$\mathcal{G}_t = (\mathcal{V}_t, \mathcal{E}_t)$ &
  map-semantic scene graph (\S\ref{sec:scene}) \\
$\rho$;\ $\sigma^{v}_t$ &
  rendering function; per-entity scene text
  $\sigma^{v}_t = \rho(v; \mathcal{G}_t)$ \\
$x_t = (m_t, g_t, u_t, \alpha_t)$ &
  bounded digest: metrics, goal progress, expansion summary, threat
  alerts (Eq.~\ref{eq:digest}) \\
$\mathcal{T}$;\ $(\tau_j, q_j, r_j)$ &
  on-demand tool set; $j$-th tool call, arguments, and result
  (Eq.~\ref{eq:react}) \\
$\mathcal{D}_t$ &
  structured plan, five schema-disjoint fields (Eq.~\ref{eq:plan}) \\
$\delta^{d}_t$ &
  directive routed to domain $d$ (Eq.~\ref{eq:routing}) \\
$\Gamma_t$ &
  cycle-level goals over five strategic domains (Eq.~\ref{eq:goal}) \\
$z_t$ &
  cycle summary with per-goal verdicts (Eq.~\ref{eq:sum}) \\
$\eta_t$ & rule-based emergency flag (\S\ref{sec:feedback}) \\
$a^{v}_t$;\ $\mathcal{A}^{v}(o_t)$ &
  action of entity $v$; its legal action set (Eq.~\ref{eq:controller}) \\
$\omega^{v}_t$;\ $h^{v}_t$ &
  entity observation slice; its recent action history \\
$\mathcal{L}_{a:b}$ &
  persisted log of digests and decisions over turns $a$--$b$ \\
$\mathcal{R}_i$;\ $A_i$ &
  episode record; post-mortem report (Eq.~\ref{eq:post}) \\
$P_i$ & strategic prior for game $i$ (Eq.~\ref{eq:evolve}) \\
$K{=}6$;\ $R{=}5$;\ $w{=}4$ &
  cycle length; controller retry cap; post-mortem window \\
$\pi_r$ &
  expert-designed role prompt encoding human strategic priors for role
  $r$ (\S\ref{sec:method}; templates in Appendix~\ref{app:prompts}) \\
$\mathrm{LLM}_{r}$ &
  backbone call under role $r$ with prompt $\pi_r$, decoded into $r$'s
  typed schema \\
\bottomrule
\end{tabular}
\end{table}

\section{Interactive Environment Comparison}
\label{app:env_comparison}

\begin{table*}[t]
\caption{%
  \textbf{Comparison of interactive environments.}
  CivRealm is the only environment satisfying all ten criteria simultaneously.
}
\label{tab:env_comparison}
\centering
\small
\renewcommand{\arraystretch}{1.1}
\setlength{\tabcolsep}{3.5pt}
\resizebox{\textwidth}{!}{%
\begin{tabular}{l cccccccccc}
\toprule
\textbf{Environment} &
  \textbf{Imperfect info} &
  \textbf{Stochastic} &
  \textbf{Multi-goal} &
  \textbf{Dynamic space} &
  \textbf{Multi-agent} &
  \textbf{General-sum} &
  \textbf{Changing players} &
  \textbf{Comm.} &
  \textbf{Tensor \& Lang.} &
  \textbf{Long-Horizon} \\
\midrule
Hanabi~\cite{hanabi2020}
  & \cmark & \xmark & \cmark & \xmark & \cmark & \cmark & \xmark & \cmark & \xmark & \xmark \\
Diplomacy~\cite{diplomacy2019}
  & \xmark & \xmark & \xmark & \xmark & \cmark & \xmark & \cmark & \cmark & \cmark & \cmark \\
Melting Pot~\cite{meltingpot2021}
  & \cmark & \cmark & \cmark & \xmark & \cmark & \cmark & \cmark & \xmark & \xmark & \xmark \\
Dota~2~\cite{dota2berner2019}
  & \cmark & \cmark & \xmark & \cmark & \cmark & \xmark & \xmark & \xmark & \xmark & \cmark \\
StarCraft~II~\cite{alphastar2019}
  & \cmark & \xmark & \xmark & \cmark & \cmark & \xmark & \xmark & \xmark & \xmark & \cmark \\
MineDojo~\cite{minedojo2022}
  & \cmark & \cmark & \cmark & \cmark & \xmark & \cmark & \xmark & \xmark & \cmark & \cmark \\
\midrule
WebArena~\cite{webarena2024}
  & \cmark & \cmark & \cmark & \cmark & \xmark & \xmark & \xmark & \xmark & \xmark & \xmark \\
OSWorld~\cite{osworld2024}
  & \cmark & \cmark & \cmark & \cmark & \xmark & \xmark & \xmark & \xmark & \xmark & \xmark \\
GAMA-Bench~\cite{gamabench2025}
  & \cmark & \cmark & \xmark & \xmark & \cmark & \cmark & \xmark & \cmark & \xmark & \xmark \\
TheAgentCompany~\cite{theagentcompany2025}
  & \cmark & \cmark & \cmark & \xmark & \cmark & \xmark & \xmark & \cmark & \cmark & \cmark \\
\midrule
\textbf{CivRealm~\cite{civrealm2024}}
  & \cmark & \cmark & \cmark & \cmark & \cmark & \cmark & \cmark & \cmark & \cmark & \cmark \\
\bottomrule
\end{tabular}
}
\end{table*}

Table~\ref{tab:env_comparison} provides a structured comparison of representative
interactive decision-making environments across ten criteria relevant to complex
strategy research. CivRealm is the only environment satisfying all ten simultaneously,
motivating its use as our primary benchmark.


\section{Experimental Setup Details \& Hyperparameters}
\label{app:hyperparameters}

In this section, we provide detailed justifications for the core hyperparameters governing the evaluation protocol of the \methodname{} framework.

\paragraph{Maximum 150 Turns ($T=150$).}
We truncate all independent runs at turn 150. In FreeCiv, the early-to-mid game (turns 1-150) represents the most critical and strategically dense phase of civilization development, requiring agents to navigate exploration, initial city foundation, technology bootstrapping, and early military encounters. By turn 150, the structural foundation of the civilization is firmly established. Evaluating beyond 150 turns heavily subjects the agent to the snowball effect (where early advantages exponentially dictate late-game success) and substantially increases inference cost without providing proportionally novel insights into the planner's fundamental reasoning capabilities.

\paragraph{Retrieval $K_r=6$.}
For the cross-game strategic evolution module, we set the retrieval budget to $K_r=6$ past experiences. Through preliminary ablation, we found that $K_r<4$ provides insufficient strategic diversity, often failing to retrieve relevant counter-strategies for specific map seeds or enemy behaviors. Conversely, $K_r>8$ exceeds the effective context window of the reasoning LLM (triggering the ``lost in the middle'' phenomenon) and dilutes the semantic weight of the most critical experiences. $K_r=6$ strikes the optimal balance between historical coverage and context tractability.

\paragraph{10 Independent Runs.}
To rigorously evaluate the baseline performance, we expand our evaluation up to 10 independent trials per algorithm on the same map seed. Strategy games like CivRealm contain significant inherent stochasticity (e.g., barbarian spawns, randomized combat outcomes, and enemy AI behavioral variance). Evaluating only 3-5 runs can lead to high variance in the reported metrics. A 10-run protocol---the largest budget-feasible under CivRealm's high per-game cost---substantially suppresses single-episode noise, supporting our reading that \methodname{}'s consistent mean lead reflects strategic capability rather than stochastic luck, even where the per-metric gap against the two strongest baselines remains within the environment's intrinsic variance.

\section{Evaluation Metric Definitions}
\label{app:metrics}

\begin{table*}[t]
\caption{\textbf{Evaluation metric definitions.} All values are snapshot readings at the end of turn 150.
  Diplomacy is a custom milestone metric and Token Consumption is measured on the LLM side; the remaining five derive from FreeCiv game internals.
  $\uparrow$ = higher is better, $\downarrow$ = lower is better.}
\label{tab:metrics_def}
\centering\footnotesize
\renewcommand{\arraystretch}{1.2}
\begin{tabularx}{\textwidth}{llX}
\toprule
\textbf{Metric} & \textbf{Dir.} & \textbf{Definition} \\
\midrule
\textbf{Score} & $\uparrow$ &
  FreeCiv built-in composite score at the end of turn 150, computed as the weighted sum of
  total population across all surviving cities, number of technologies researched, and
  accumulated Wonder scores.  It serves as the primary holistic measure of civilizational
  development. \\
\textbf{Technology} & $\uparrow$ &
  Number of distinct technologies researched by turn 150.
  Each technology is counted once regardless of how many cities benefit. \\
\textbf{Military Power} & $\uparrow$ &
  Sum of the attack-strength values of all surviving owned military units at the end of
  turn 150.  Units destroyed in combat are not counted; only units alive at episode
  termination are included, weighted by their base attack rating in the FreeCiv classic
  ruleset.  A stronger military force indicates a greater capacity to defend the civilization
  against enemy raids and protect productive cities from destruction. \\
\textbf{Cities} & $\uparrow$ &
  Number of cities still owned at the end of turn 150.
  Cities captured or destroyed before turn 150 are not counted.
  City count is a proxy for population potential: each city independently produces food,
  production, and science, so a larger city base translates directly to faster growth,
  higher research output, and greater resource accumulation in subsequent turns. \\
\textbf{Buildings} & $\uparrow$ &
  Total number of buildings present across all surviving owned cities at the end of
  turn 150.  Buildings in cities that were captured or razed before turn 150
  are not included; World Wonders count as one building each.
  Buildings provide compounding bonuses across all dimensions of civilizational
  development---Granaries accelerate population growth, Libraries and Marketplaces
  boost science and gold output, and Barracks improve military unit quality---making
  total building count a reliable indicator of long-term infrastructure investment. \\
\textbf{Diplomacy} & $\uparrow$ &
  Custom milestone score reflecting de-escalation progress with each AI opponent.
  Points are awarded cumulatively for each milestone reached: Ceasefire~(+5 pts), Armistice~(+10 pts),
  Peace~(+15 pts), and Alliance~(+20 pts), for a maximum of 50 pts per AI opponent.
  This metric is not provided by the FreeCiv engine and is computed by post-processing the diplomatic-state log. \\
\textbf{Token Consumption} & $\downarrow$ &
  Total LLM input and output token counts over 150 turns, normalized to a single episode.
  Detailed counting methodology per method is given in Appendix~\ref{sec:app_token}. \\
\bottomrule
\end{tabularx}
\end{table*}

All seven metrics are evaluated at turn 150 of each episode.
Table~\ref{tab:metrics_def} provides precise definitions; the first six capture civilization
development breadth, while the seventh measures computational cost.

\paragraph{Significance of Core Metrics.}
\textbf{Score} is the holistic measure of civilizational capability and is our primary outcome. Among the component metrics, \textbf{Buildings} is the most informative differentiator: it captures the infrastructure investment that compounds across growth, research, and defense, is scored independently of Score, and is the one dimension on which \methodname{} significantly surpasses \emph{every} baseline (Appendix~\ref{app:stats}). \textbf{Military Power} reflects survivability but is extremely high-variance under CivRealm's stochastic combat and barbarian raids (its standard deviation often exceeds its mean), and \textbf{Diplomacy} is largely exploration-driven; we therefore report both for completeness but do not rest our central claims on them. The Score formula aggregates population across cities, researched technologies, and Wonder scores---so Technology and city-driven population are partly coupled to Score, whereas Buildings and Diplomacy are scored independently.

\FloatBarrier
\section{Distribution-Aware Statistics and Significance Tests}
\label{app:stats}

Because CivRealm scores are heavy-tailed---a single fortunate or catastrophic
episode shifts the mean substantially---we complement the $\mathrm{mean}\pm
\mathrm{std}$ of Table~\ref{tab:main} with rank-based and distribution-aware
statistics. For each metric we run a two-sided Mann--Whitney $U$ test of
\methodname{} against each baseline and report the median with inter-quartile
range $[\mathrm{P25},\mathrm{P75}]$ and the head-to-head win-rate---the
probability that a random \methodname{} game outscores a random baseline game,
i.e.\ the common-language effect size (Table~\ref{tab:stats}).

\begin{table*}[t]
  \caption{\textbf{Distribution-aware statistics and pairwise significance} for the
    two headline metrics (Map Seed 2025; $n{=}10$ per method).
    For each baseline we report the two-sided Mann--Whitney $p$ against
    \methodname{} and the head-to-head win-rate.
    Medians with $[\mathrm{P25},\mathrm{P75}]$ complement the means of
    Table~\ref{tab:main} under CivRealm's heavy-tailed distribution.
    \methodname{} significantly leads \emph{all} baselines on Buildings; on Score
    it significantly leads the weaker three but lies within noise against CoS and
    HIMA. Significant cells ($p<0.05$) in \textbf{bold}.}
  \label{tab:stats}
  \centering
  \footnotesize
  \setlength{\tabcolsep}{6pt}
  \begin{tabular}{l c c c c c c c}
    \toprule
    & & \multicolumn{3}{c}{\textbf{Score}} & \multicolumn{3}{c}{\textbf{Buildings}} \\
    \cmidrule(lr){3-5}\cmidrule(lr){6-8}
    \textbf{Method} & $n$ &
      Med.~[P25,\,P75] & $p$ vs Ours & Win\% &
      Med.~[P25,\,P75] & $p$ vs Ours & Win\% \\
    \midrule
    \textbf{\methodname{} (Ours)} & 10 & 55.5~[50.0,\,71.5] & --- & --- & 11.5~[8.8,\,19.2] & --- & --- \\
    \midrule
    CoS       & 10 & 42.5~[34.0,\,55.8] & 0.064 & 75\% & 5.5~[2.5,\,11.2] & \textbf{0.037} & 78\% \\
    EpicStar  & 10 & 28.0~[23.8,\,33.5] & \textbf{0.003} & 90\% & 2.0~[1.0,\,4.0] & \textbf{0.001} & 92\% \\
    Optimus-2 & 10 & 40.5~[38.0,\,44.8] & \textbf{0.004} & 89\% & 1.0~[1.0,\,2.0] & \textbf{$<$0.001} & 98\% \\
    HIMA      & 10 & 36.5~[27.2,\,55.8] & 0.096 & 72\% & 2.5~[1.2,\,4.5] & \textbf{$<$0.001} & 97\% \\
    Mastaba   & 10 & 40.0~[27.5,\,48.8] & \textbf{0.007} & 86\% & 1.0~[1.0,\,2.0] & \textbf{$<$0.001} & 97\% \\
    \bottomrule
  \end{tabular}
\end{table*}

\paragraph{Where \methodname{}'s advantage is significant.}
Summarizing the per-metric Mann--Whitney tests against the five baselines
(significant at $p<0.05$): \textbf{Buildings}, $5/5$ baselines---the only metric
that survives Holm correction within its per-metric family, and $4/5$ under the
stricter $30$-comparison family-wise correction (the CoS comparison, $p{=}0.037$,
becomes suggestive); \textbf{Cities},
$4/5$ (all but HIMA); \textbf{Score}, $3/5$ (not significant against CoS and
HIMA); \textbf{Military}, $3/5$ (CoS, EpicStar, Mastaba); \textbf{Diplomacy},
$3/5$ (not significant against CoS and EpicStar, consistent with its
exploration-driven variance noted in Appendix~\ref{app:metrics}); and
\textbf{Technology}, $1/5$ (only EpicStar; tied with the four stronger
baselines). Infrastructure is therefore the
statistically robust core of \methodname{}'s improvement; its Score advantage is
large and consistent in the mean but, against the two strongest baselines, lies
within the environment's intrinsic noise---which we state plainly rather than
obscure behind aggregate means. With five baselines and six metrics ($30$
comparisons), individual $p$-values near $0.05$ should be read as suggestive;
the Buildings result is the one that best survives family-wise correction,
remaining significant against four of five baselines even under the full
$30$-comparison family.

\FloatBarrier
\section{Algorithm Framework Pseudocode}
\label{app:algorithm}

Algorithm~\ref{alg:saga} provides the formal pseudocode for the
\methodname{} framework.
Each line that invokes the LLM backbone is annotated with the
corresponding equation from \S\ref{sec:method}, making explicit which
role is called, on which inputs, and what structured output it returns;
all remaining lines are deterministic rules.

\begin{algorithm}[!htbp]
\caption{\methodname{} Framework Execution}
\label{alg:saga}
\begin{algorithmic}[1]
\REQUIRE $N$ games, max turns $T_{\max}$, cycle length $K{=}6$,
  retry cap $R{=}5$, post-mortem window $w{=}4$
\STATE $P_1 \leftarrow \emptyset$;\quad $A_0 \leftarrow \emptyset$
  \COMMENT{strategic prior, post-mortem}
\FOR{game $i = 1$ \TO $N$}
  \STATE load $A_{i-1}$ and $P_i$ from disk; $\Gamma \leftarrow
    \emptyset$; $z \leftarrow \emptyset$; log $\mathcal{L} \leftarrow
    \emptyset$
  \FOR{turn $t = 1, 2, \ldots$ \textbf{while} game not over \AND $t
    \le T_{\max}$}
    \STATE $\mathcal{G}_t \leftarrow \mathrm{BuildGraph}(o_t)$;\quad
      $\sigma^{v}_t \leftarrow \rho(v; \mathcal{G}_t)$ for each entity
      $v$ \COMMENT{\S\ref{sec:scene}, rule-based}
    \STATE $x_t \leftarrow \mathrm{digest}(o_t, \mathcal{G}_t,
      \Gamma)$ \COMMENT{Eq.~\ref{eq:digest}}
    \STATE $\eta_t \leftarrow$ rule-based emergency predicate on $o_t$
      \COMMENT{\S\ref{sec:feedback}}
    \IF{$t \bmod K = 1$ \OR $\eta_t = 1$}
      \STATE $z \leftarrow \mathrm{LLM}_{\mathrm{sum}}(\pi_{\mathrm{sum}}; \Gamma,
        \mathcal{L}_{t-K:t}, x_t)$ \COMMENT{Eq.~\ref{eq:sum}: per-domain
        goal verdicts}
      \STATE $\Gamma \leftarrow \mathrm{LLM}_{\mathrm{goal}}(\pi_{\mathrm{goal}}; x_t, z,
        A_{i-1}, P_i, \eta_t)$ \COMMENT{Eq.~\ref{eq:goal}: posture +
        per-domain goals}
    \ENDIF
    \STATE $\mathcal{D}_t \leftarrow$ ReAct loop of
      $\mathrm{LLM}_{\mathrm{plan}}(\pi_{\mathrm{plan}}; x_t, \bar\sigma_t, \Gamma, z,
      \cdot)$ with tools $\mathcal{T}$
      \COMMENT{Eqs.~\ref{eq:react}--\ref{eq:plan}}
    \STATE $\delta^{d}_t \leftarrow$ route $(\mathcal{D}_t, \Gamma)$
      per domain $d$ \COMMENT{Eq.~\ref{eq:routing}}
    \FORALL{active entities $v$, \textbf{in parallel} (diplomacy
      serialized)}
      \STATE $a^{v}_t \leftarrow \mathrm{LLM}_{d(v)}(\pi_{d(v)}; \delta^{d(v)}_t,
        \sigma^{v}_t, \omega^{v}_t, h^{v}_t)$
        \COMMENT{Eq.~\ref{eq:controller}}
      \STATE validate $a^{v}_t \in \mathcal{A}^{v}(o_t)$; on violation
        re-prompt ($\le R$), else no-op
    \ENDFOR
    \STATE execute $\{a^{v}_t\}$ in the environment; append digests
      and decisions to $\mathcal{L}$
  \ENDFOR
  \STATE $A_i \leftarrow \mathrm{LLM}_{\mathrm{post}}(\pi_{\mathrm{post}}; \mathcal{R}_i)$
    \COMMENT{Eq.~\ref{eq:post}: post-mortem from episode record}
  \STATE $P_{i+1} \leftarrow \mathrm{LLM}_{\mathrm{evolve}}(\pi_{\mathrm{evolve}}; A_{i-w+1},
    \ldots, A_i)$ \COMMENT{Eq.~\ref{eq:evolve}: windowed prior rebuild}
\ENDFOR
\end{algorithmic}
\end{algorithm}

\section{Capability Comparison}
\label{app:comparison}

In this appendix, we provide a detailed qualitative comparison of various LLM-based strategy agents. As the complexity of environments like CivRealm scales, different architectural paradigms have emerged to handle vast state spaces, long-horizon planning, and cross-episode strategic evolution. Table~\ref{tab:comparison} outlines these distinctions, contrasting our approach against state-of-the-art baselines across five core design dimensions.

\begin{table*}[t]
\caption{%
  \textbf{Qualitative paradigm comparison} of LLM strategy agents across five
  design dimensions. Each cell names the concrete mechanism employed, enabling
  direct comparison of architectural choices rather than binary feature presence.
}
\label{tab:comparison}
\centering
\small
\renewcommand{\arraystretch}{1.2}
\resizebox{\textwidth}{!}{%
\begin{tabular}{llllll}
\toprule
\textbf{Method} &
  \textbf{Input State} &
  \textbf{Action Space} &
  \textbf{Planning Horizon} &
  \textbf{Cross-Game Learn.} &
  \textbf{Action Constraints} \\
\midrule
CoS~\cite{cos2024}
  & Rolling summary & Monolithic output & Per-turn reactive & Trace replay & Prompt-based \\
EpicStar~\cite{epicstar2025}
  & Episodic retrieval & Monolithic output & Per-turn reactive & Episodic RAG & Prompt-based \\
Optimus-2~\cite{optimus2}
  & Full-state injection & Dual-track sequence & Phase milestones & Trace replay & Prompt-based \\
HIMA~\cite{hima2025}
  & Hierarchical filtering & 3-advisor council & Per-turn reactive & Static profiling & Prompt-based \\
Mastaba~\cite{civrealm2024}
  & Pyramid hawk-eye & Advisor + workers & Per-turn reactive & Manual + hist.\ RAG & API list + prompt \\
\midrule
\textbf{\methodname{} (Ours)}
  & \textbf{Graph + Tool pull} & \textbf{6-way decoupled} & \textbf{Periodic + Interrupt} & \textbf{Causal distillation} & \textbf{Schema-constrained} \\
\bottomrule
\end{tabular}%
}
\\[5pt]
\raggedright\footnotesize
\textit{Key distinctions.}
\textbf{State:} CoS compresses turn history; EpicStar retrieves similar past episodes;
Mastaba condenses the map into a pyramid hawk-eye view with per-unit tactical zoom;
\methodname{} builds a relational graph and fetches numeric data only on demand---avoiding
both stale context and token inflation.
\textbf{Action \& Horizon:} Baselines emit a single unified plan per turn or delegate
via advisor hierarchies; \methodname{} routes decisions to six fully independent
specialist controllers (five concurrent, diplomacy serialized) and enforces a separate periodic subgoal loop with interrupt.
\textbf{Learning \& Constraints:} Unlike trace replay, semantic RAG, or manual-document
retrieval (Mastaba), our causal distillation extracts transferable strategic rules.
Environment-level action-mask validation is shared infrastructure across \emph{all}
methods; \methodname{}'s distinction is that its structured output schema constrains the
Planner to currently-legal actions (e.g., city production restricted to
$\mathcal{A}^{city}(\tau_t)$), reducing illegal proposals upstream of the mask rather than
relying on prompt-injected behavioral warnings.
\end{table*}

\FloatBarrier
\section{Token Consumption Analysis}
\label{sec:app_token}

This section details the specific computation mechanics for the token consumption analysis reported in Table~\ref{tab:token}. To avoid double counting and maintain fairness across structurally diverse agent frameworks, we adhere to the following normalization principles:

\begin{table}[t]
  \caption{\textbf{Token consumption breakdown} per method, normalised to 150 turns
    (mean\,$\pm$\,std over independent runs). $\downarrow$ = lower is better.}
  \label{tab:token}
  \centering
  \footnotesize
  \setlength{\tabcolsep}{4pt}
  \begin{tabular}{lccc}
    \toprule
    \textbf{Method} &
      \textbf{Input}\,$\downarrow$ &
      \textbf{Output}\,$\downarrow$ &
      \textbf{Total}\,$\downarrow$ \\
    \midrule
    CoS       & $132{\pm}46$  & $149{\pm}21$  & $281{\pm}51$ \\
    EpicStar  & $213{\pm}140$ & $58{\pm}4$    & $270{\pm}140$ \\
    Optimus-2 & $246{\pm}97$  & $66{\pm}7$    & $312{\pm}97$ \\
    HIMA      & $168{\pm}94$  & $219{\pm}43$  & $387{\pm}103$ \\
    Mastaba   & $14{\pm}1$    & $62{\pm}26$   & $76{\pm}26$ \\
    \midrule
    \textbf{\methodname{} (Ours)}
              & $258{\pm}54$  & $109{\pm}14$  & $367{\pm}56$ \\
    \bottomrule
  \end{tabular}
  \\[3pt]
  {\footnotesize All counts in thousands (k) of tokens.}
\end{table}

\begin{itemize}
    \item \textbf{Input Tokens}: Represents all tokens strictly parsed or embedded within the final user/system prompts sent to the decision-making model.
    \item \textbf{Output Tokens}: Represents all raw generation tokens emitted by the LLMs acting as the main decision interface.
\end{itemize}

\paragraph{Method-Specific Calculation Logic.}
For \textbf{CoS}, all observations and historical rolling summaries are computed as Input.
\textbf{EpicStar} employs a local TF-IDF and heuristic working memory; since these memories are constructed locally without LLM inference, their text is entirely attributed to the Input prompt without inflating the Output cost.
\textbf{Optimus-2} utilizes a sub-LLM to summarize behaviors, and to fairly portray the system-level cost, these generated sub-goals and behavioral summaries are counted as Input when fed to the main action loop.
\textbf{HIMA} operates a multi-agent council system where three advisors and one synthesizer interact in parallel; the aggregate of all advisor outputs forms a substantial portion of the output token cost per turn.
\textbf{Mastaba} builds an extremely compact global state sketch (turn number, unit/city counts, war-state flag, and economic rates) from raw observations using a rule-based aggregator, bypassing any LLM-generated intermediate representation. This sketch typically contains fewer than 150 tokens, which explains its ultra-low input token count ($14\text{k}{\pm}1\text{k}$) compared to all other baselines.
Finally, for \textbf{\methodname{} (Ours)}, we directly calculate the input and output tokens by summing the usage across the \texttt{SetGoalAgent}, \texttt{AbstractAgent}, and the central Planner. By adopting this holistic accounting, we capture the true system-level cost of our hierarchical design, demonstrating that while the auxiliary \texttt{SetGoal} and \texttt{Abstract} agents add modest token overhead, they drastically reduce the Planner's output burden by offloading complex strategic reasoning.

\section{Backbone Generalization}
\label{sec:backbone}

\begin{table}[t]
  \caption{\textbf{Backbone generalization.} The same single-game protocol as
    Table~\ref{tab:main} (Map Seed 2025, mean\,$\pm$\,std over five 150-turn
    games, base planner only), but with the backbone LLM swapped from the strong
    Doubao-Seed-1.8 to the markedly weaker GPT-4o-mini. All observation wrappers,
    controllers, and tools are held fixed; only the backbone changes.
    \methodname{} leads on all six metrics. Best per metric in \textbf{bold}.}
  \label{tab:backbone}
  \centering
  \footnotesize
  \setlength{\tabcolsep}{5pt}
  \begin{tabular}{lccc}
    \toprule
    \textbf{Metric} & \textbf{CoS} & \textbf{HIMA} &
      \textbf{\methodname{} (Ours)} \\
    \midrule
    Score\,$\uparrow$      & 28.8\,$\pm$\,16.3 & 27.4\,$\pm$\,15.0 & \textbf{51.2\,$\pm$\,19.5} \\
    Tech\,$\uparrow$       & 4.6\,$\pm$\,0.5   & 4.8\,$\pm$\,1.2   & \textbf{11.4\,$\pm$\,1.6} \\
    Military\,$\uparrow$   & 21.4\,$\pm$\,25.6 & 26.2\,$\pm$\,26.9 & \textbf{39.6\,$\pm$\,21.0} \\
    Cities\,$\uparrow$     & 6.6\,$\pm$\,7.7   & 7.0\,$\pm$\,6.9   & \textbf{9.8\,$\pm$\,6.6} \\
    Buildings\,$\uparrow$  & 2.0\,$\pm$\,2.3   & 1.6\,$\pm$\,1.5   & \textbf{6.8\,$\pm$\,3.8} \\
    Diplomacy\,$\uparrow$  & 0.0\,$\pm$\,0.0   & 0.0\,$\pm$\,0.0   & \textbf{10.0\,$\pm$\,20.0} \\
    \bottomrule
  \end{tabular}
\end{table}

A natural concern is whether \methodname{}'s advantage is tied to a single
strong backbone LLM. To test this, we replace the Doubao-Seed-1.8 backbone of
Table~\ref{tab:main} with the markedly weaker GPT-4o-mini, hold every
observation wrapper, controller, and tool fixed, and re-run the identical
single-game protocol (Map Seed 2025, five 150-turn games, base planner only)
against \methodname{}'s two strongest baselines by Score, CoS and HIMA (the other
three, already weaker under the strong backbone in Table~\ref{tab:main}, are omitted
under budget constraints; Table~\ref{tab:backbone}).

Under this weaker backbone, \methodname{} still wins \emph{every} metric. It
leads the stronger baseline (CoS) by $78\%$ on Score ($51.2$ vs.\ $28.8$),
attains roughly $2.4\times$ the Tech and over $3\times$ the Buildings of either
baseline, and simultaneously secures the most Cities and the strongest Military.
The capability ordering of Table~\ref{tab:main} therefore survives a substantial
downgrade of the reasoning engine: structured scene-aware perception and
goal-directed execution help \emph{regardless} of backbone strength, rather than
merely amplifying an already-capable LLM.

As expected, absolute performance drops relative to Doubao
(\methodname{} Score $60.7\!\to\!51.2$), and the large Military variance
($39.6\pm21.0$) reflects mid-game territory losses to pirate and AI incursions
that a small model recovers from unevenly. Diplomacy nearly vanishes for all
three methods---only \methodname{} brokers a single alliance across the fifteen
games---suggesting that multi-party treaty formation is the first capability to
exceed a weak backbone's reasoning ceiling. We examine that ceiling directly in
Appendix~\ref{app:case_seed2029}: a cross-game \emph{evolution} case study on a
deliberately resource-poor map (Map Seed 2029) shows that even a correct
distilled strategy delivers only muted gains once the backbone---not the
strategy---becomes the binding constraint on decision quality.

\FloatBarrier
\section{Single-Game Failure Modes (Map Seed 2025)}
\label{app:failure_cases}

Our two lowest-scoring Seed-2025 episodes make the single-game
\emph{overcorrection} discussed in \S\ref{sec:analysis} concrete.
In the first, core cities looped on redundant units, starving
marketplaces and wonders, while frontier cities received garrisons but
never walls and were lost late-game.
In the second, the empire froze into a defensive posture under
sustained barbarian pressure, holding science at $20\%$, stalling its
technology chain, and slipping from Monarchy back to Despotism.
In both cases the post-game analyst diagnosed the imbalance and
injected corrective priors (e.g., \emph{core cities must prioritize
infrastructure}) into the next episode, within the limits set by
backbone capability (Appendix~\ref{app:case_seed2029}).

\section{Cross-Backbone Evolution Case Study (Map Seed 2029)}
\label{app:case_seed2029}

\begin{figure}[t]
  \centering
  \includegraphics[width=\columnwidth]{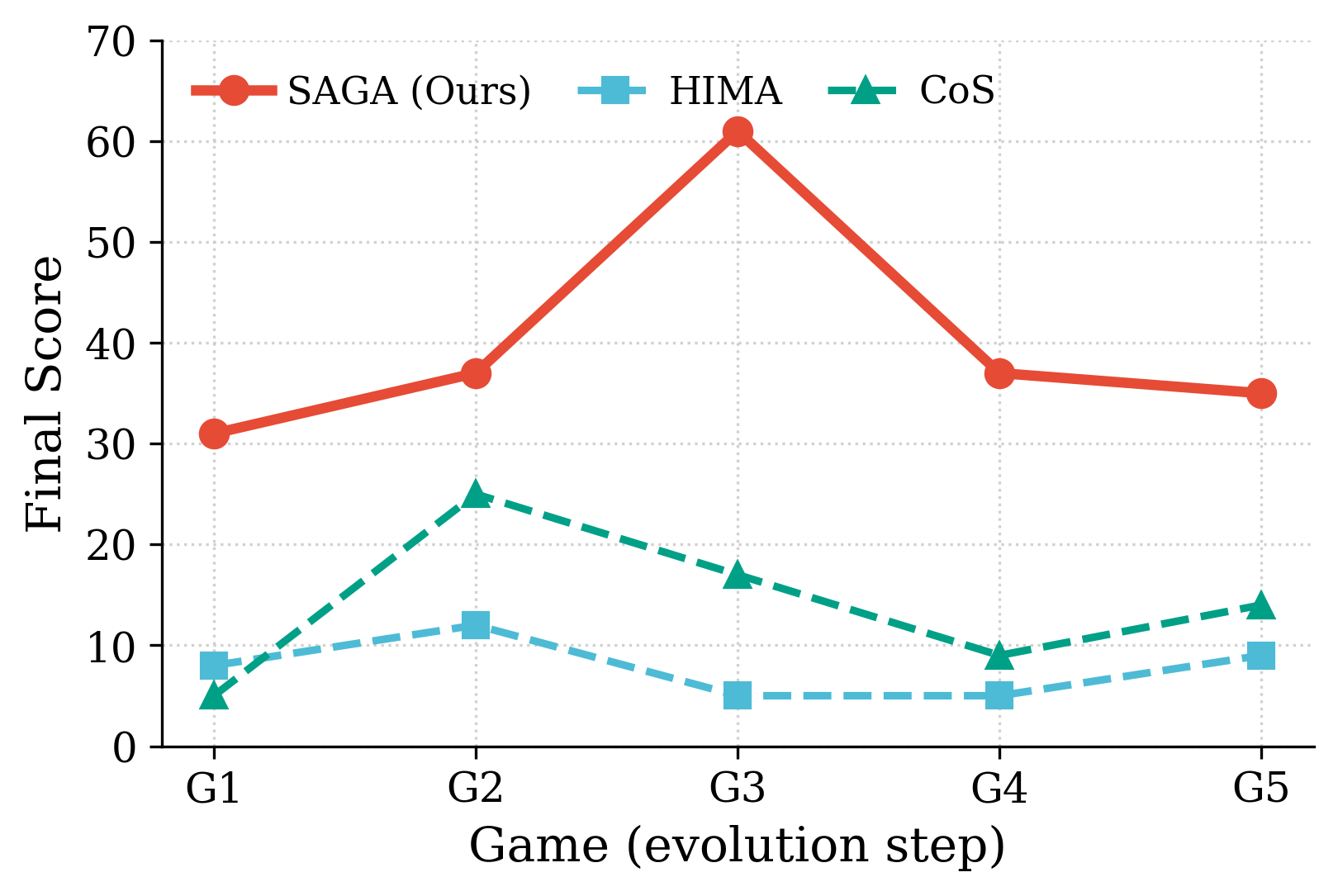}
  \caption{\textbf{GPT-4o-mini evolution trajectory on Map Seed 2029.}
    Final Score across the five-game evolution chain (G1$\to$G5) on the
    resource-poor map. \methodname{} (solid red) dominates both baselines at
    every step and peaks at G3 (Score~$61$) before regressing as it lapses back
    into Despotism. The weak backbone makes gains \emph{non-monotonic}, yet
    \methodname{} never drops into the oscillating low band occupied by HIMA and
    CoS---structured execution keeps it consistently ahead even when evolution's
    headroom is throttled.}
  \label{fig:gpt_evolution}
\end{figure}

This case study supports the backbone-generalization analysis of
Appendix~\ref{sec:backbone} by asking a sharper question: when the backbone itself is
weak, does cross-game \emph{evolution} still pay off? We run a five-game
evolution chain (G1$\to$G5) under GPT-4o-mini on \emph{Map Seed 2029}---a
deliberately harsh, resource-poor map dominated by desert, ocean, and swamp
tiles that sharply restricts viable city sites---in stark contrast to the
resource-rich Map Seed 2025 used throughout the main results. The full evolution
loop is enabled (\texttt{GameAnalyst} post-mortem $+$ injected pre-game
strategic prior); all three methods share identical observation, controller,
tool, and evolution-chain infrastructure, so only the planner layer differs.
Table~\ref{tab:case2029} reports the per-game Final Score and surviving city
count for each method, and Figure~\ref{fig:gpt_evolution} plots the
corresponding Score trajectory.

\begin{table}[t]
  \caption{\textbf{Cross-backbone evolution on the resource-poor Map Seed 2029}
    (GPT-4o-mini, evolution loop enabled, Final values per game). \methodname{}
    records the highest Score and the largest surviving territory in every game
    and never collapses, whereas both baselines repeatedly fall to zero cities.
    Best per game in \textbf{bold}.}
  \label{tab:case2029}
  \centering
  \footnotesize
  \setlength{\tabcolsep}{6pt}
  \begin{tabular}{lccccc}
    \toprule
    \textbf{Method} & \textbf{G1} & \textbf{G2} & \textbf{G3} & \textbf{G4} & \textbf{G5} \\
    \midrule
    \multicolumn{6}{l}{\emph{Score} (Final)} \\
    \quad\methodname{} & \textbf{31} & \textbf{37} & \textbf{61} & \textbf{37} & \textbf{35} \\
    \quad HIMA         & 8  & 12 & 5  & 5  & 9  \\
    \quad CoS          & 5  & 25 & 17 & 9  & 14 \\
    \midrule
    \multicolumn{6}{l}{\emph{City count} (Final)} \\
    \quad\methodname{} & \textbf{5} & \textbf{8} & \textbf{7} & \textbf{7} & \textbf{9} \\
    \quad HIMA         & 0 & 3 & 1 & 1 & 4 \\
    \quad CoS          & 0 & 6 & 0 & 0 & 2 \\
    \bottomrule
  \end{tabular}
\end{table}

Two findings stand out. First, even with the evolution loop active, the gains
are \emph{non-monotonic} and muted. \methodname{} peaks at G3 (Score $61$: a
$14$-tech run that transitions into Monarchy and erects nineteen buildings) but
regresses to $37$/$35$ at G4/G5, where it lapses back into Despotism and
over-militarizes at the expense of technology and construction (Tech
$14\!\to\!8\!\to\!5$). The post-mortem correctly diagnoses this drift each
episode and injects a sound corrective prior, yet the GPT-4o-mini planner cannot
reliably act on it---a direct illustration that a weak backbone's
\emph{decision ceiling}, rather than the quality of the distilled advice,
becomes the binding constraint. Even a correct strategy helps only as much as
the executor can carry it out.

Second, this ceiling sits far lower on a resource-poor map. Both absolute scores
and inter-game improvements are markedly smaller than under the resource-rich
Seed 2025 main experiment, and the two baselines repeatedly collapse to zero
surviving cities (CoS at G1/G3/G4, HIMA at G1). The scarce terrain leaves almost
no margin for the recovery maneuvers that a stronger backbone would execute
automatically.

Despite these limits, the \emph{relative} picture is unchanged: \methodname{}
records the highest Score and the largest surviving territory in every game and
never suffers civilizational collapse, whereas the baselines oscillate between
marginal survival and total wipe-out. Structured execution thus remains
advantageous even when evolution's headroom is throttled by a weak
backbone---it is precisely what keeps the agent alive long enough for any
strategic prior to matter at all.

\FloatBarrier
\section{Implementation Details}
\label{app:impl}

\subsection{Shared Infrastructure}

All methods share the same LLM backbone (Doubao-Seed-1.8) and execution infrastructure.
The Planner and all entity controllers (City, Unit, Gov, Tech, Diplomacy) are built on a
unified \texttt{ReActAgent} wrapper that writes action logs to per-run SQLite databases
in WAL mode, enabling concurrent multi-threaded reads without contention.

All tool-calling planners (SAGA, CoS, HIMA, Optimus-2) apply a maximum of \textbf{5 sequential
tool-call iterations} per planning step, after which a forced-finalization message is injected to
obtain a valid JSON output. This prevents infinite tool-call loops on complex states while
still allowing multi-step information gathering.
All controllers validate generated actions against the environment's \texttt{available\_actions}
mask before execution; illegal actions trigger an inline retry without advancing the game clock,
ensuring the comparison is not confounded by action legality failures.

\paragraph{Inference parameters.}
All LLM agents query the same provider endpoint with a per-call output cap of
$4096$ tokens, a rolling chat-history window of $20$ messages, and up to $3$
automatic retries on transport errors. Decoding uses low-temperature,
near-deterministic sampling held identical across all methods; the exact
temperature and top-$p$ values are fixed in---and released with---our
configuration files, so the backbone and sampling regime stay constant across
the comparison.

\subsection{\methodname{} Configuration}

\begin{table}[!t]
  \caption{On-demand tools available to the \methodname{} planner.
    Tool calls replace pre-injected state dumps, reducing input tokens
    while improving decision accuracy.}
  \label{tab:tools}
  \centering
  \small
  \begin{tabularx}{\columnwidth}{@{}l >{\raggedright\arraybackslash}X@{}}
    \toprule
    \textbf{Tool} & \textbf{Returns \& Usage} \\
    \midrule
    \texttt{query\_city\_metrics} &
      Per-city production, food, gold, science, happiness, unit burden,
      and recent production history (items built over the past turns).
      \newline \textit{Used before:} City plan \\
    \addlinespace
    \texttt{query\_army\_roster} &
      Full unit table (type, pos, HP, MP, veteran, fortified) + city garrison.
      \newline \textit{Used before:} Military plan \\
    \addlinespace
    \texttt{query\_tech\_status} &
      Researched count, current research, next 3 available techs.
      \newline \textit{Used before:} Tech plan \\
    \bottomrule
  \end{tabularx}
\end{table}

\paragraph{On-demand tool suite.}
Table~\ref{tab:tools} lists the three tools available to the planner and
the plan each precedes.

\paragraph{Inner-loop scheduler.}
The \texttt{AbstractAgent} and \texttt{SetGoalAgent} are invoked every $K=6$ turns by default.
In addition, an event-driven \emph{emergency interrupt} triggers an immediate re-evaluation
when any of the following conditions is detected:
(i)~an increase in visible enemy unit count;
(ii)~enemy military units entering a city's proximity radius;
(iii)~a newly discovered civilization or player;
(iv)~the loss of a friendly city.
The interrupt is subject to a \textbf{5-turn cooldown} to prevent cascading re-evaluations that
would inflate token consumption.

\paragraph{SceneGraph topology.}
The graph is rebuilt from scratch every turn in $<$10~ms (CPU).
Beyond the standard within-radius edges, two global structural rules are applied:
(i)~every allied unit maintains a permanent long-range directed edge to its nearest friendly
city regardless of map distance, preventing Settlers from losing navigation context in
unexplored regions;
(ii)~every allied military unit maintains directed edges to all visible enemy cities as
attack-route anchors.
Edge distances are discretized into three levels---\emph{close} ($<3$ tiles), \emph{medium}
($<10$ tiles), and \emph{far} ($\geq 10$ tiles)---to reduce the cognitive load of parsing
raw coordinate differences for the LLM.

\paragraph{Concurrency and rate limiting.}
The \texttt{Thread\-Pool\-Executor} runs up to \textbf{20} workers, and a
\texttt{threading.Semaphore(20)} caps the number of simultaneously active LLM calls at
\textbf{20}, preventing API rate-limit exhaustion during heavy mid-game turns with many
concurrent city and unit controllers.

\subsection{Baseline-Specific Configuration}

\paragraph{CoS~\cite{cos2024}.}
Implements an 8-stage Chain-of-Summarization (CoS) prompting pipeline each turn. The
sliding history window retains the \textbf{most recent 1 L1-summary} (reduced from 5 in the
original paper to fit the CivRealm token budget). Planner output is capped at 3,000 tokens
with a 1,500-word hard limit enforced via prompt instruction to prevent late-game generation
timeouts.

\paragraph{EpicStar~\cite{epicstar2025}.}
The episodic memory module stores up to \textbf{200 episodes} per game and retrieves the
top-$k{=}3$ most relevant past episodes using TF-IDF cosine similarity (vocabulary capped at
\texttt{max\_features=500}). A keyword-overlap fallback is activated when \texttt{sklearn} is
unavailable. Working memory carries the current observation summary and the single most recent
planning decision.

\paragraph{Optimus-2~\cite{optimus2}.}
Maintains an active sub-goal and a prioritized sub-goal queue. Re-planning is triggered
every \textbf{5 turns} or after the active sub-goal has been held for \textbf{10 consecutive
turns} without progression, whichever comes first. The observation history window covers
\textbf{3 L1-summaries}. On non-replan turns, a lightweight passthrough response propagates
the unchanged active sub-goal to all downstream controllers, avoiding redundant LLM calls.

\paragraph{HIMA~\cite{hima2025}.}
Three parallel advisors---\emph{Expansion \& Economy}, \emph{Military \& Defense}, and
\emph{Technology \& Development}---each independently generate proposals via chain-of-thought
reasoning. A Strategic Planner (SP) then fuses the three proposals into a single unified plan.
History window: \textbf{1 L1-summary}. Total LLM calls per turn: 3 (advisors, run in parallel
via \texttt{ThreadPoolExecutor}) $+$ 1 (SP fusion) $=$ 4, which explains HIMA's elevated
output token count in Table~\ref{tab:token}.

\paragraph{Mastaba~\cite{civrealm2024}.}
Mastaba employs a flat two-level information architecture: a single \emph{Advisor} generates
a unified natural-language suggestion from a compressed global sketch, which is then
broadcast verbatim to all entity controllers (workers, military, settlers) as their sole
strategic directive. The global sketch is assembled by a rule-based aggregator that counts
unit types, city sizes, and war state---without any LLM inference---and serializes the result
into a compact string ($<$150 tokens). This design explains Mastaba's exceptionally low token
footprint (Table~\ref{tab:token}). There is no per-entity observation or context window;
all controllers receive the identical advisor broadcast, which limits strategic specificity
but reduces API cost to a single LLM call per turn. Max output tokens: \textbf{2,500};
stateful fallback re-uses the last valid suggestion if the current turn's LLM call fails.

\section{Prompt Templates and SceneGraph Rendering}
\label{app:prompts}

We present representative excerpts of the four core prompt templates used in \methodname.
Full prompts contain extensive game-mechanic rules and self-check chains; here we show only the structural skeleton and the most design-critical directives.
All prompts use a \texttt{system} + \texttt{user} two-message format and receive structured JSON output.

\subsection{Pre-Game Strategic Analyst ($\mathrm{LLM}_{\mathrm{evolve}}$, Eq.~\ref{eq:evolve})}

Invoked once before each game using all prior \texttt{game\_analysis\_report.json} files.
Generates a \texttt{strategic\_evolution\_chain.md} that summarizes cross-game lessons.

\begin{tcolorbox}[colback=gray!7, colframe=gray!50, title={\textbf{Pre-Game Analyst} — System Prompt (Condensed)}, fonttitle=\small\bfseries, left=4pt, right=4pt, top=3pt, bottom=3pt, breakable]
\small\ttfamily
You are the Chief Strategic Architect for the FreeCiv AI.\par
Perform an "Evolutionary Forensic Analysis" across historical game outcomes.\par
\vspace{3pt}
For EVERY game, you MUST provide:\par
  inherited\_strategy: "Adopted the [Strategy] from previous [Game N]"\par
  strategic\_modification: "Due to failure of [Game N] in [Reason], ..."\par
  post\_adjustment\_successes: Acknowledge measurable improvements.\par
  deep\_root\_cause: Chain-of-Failure causal trace (5 levels minimum).\par
\vspace{3pt}
upcoming\_game\_expectations: Frame as SELF-CHECK QUESTIONS, not rigid rules.\par
  e.g. "Before switching to SURVIVAL, ask: Has a city actually been lost?"\par
Output JSON: \{evolution\_chain, persistent\_failures,\par
\hspace*{1.4em}upcoming\_game\_expectations, ...\}
\end{tcolorbox}

\subsection{SetGoal Agent ($\mathrm{LLM}_{\mathrm{goal}}$, Eq.~\ref{eq:goal})}

Called every $K$ turns to set empire-level strategic mode and 5-turn sub-goals.
Receives two cross-game inputs: the \texttt{AbstractAgent}'s natural-language summary of recent
in-game events, and the \texttt{strategic\_evolution\_chain.md} compiled by the Pre-Game Analyst.

\begin{tcolorbox}[colback=blue!4, colframe=blue!40, title={\textbf{SetGoal Agent} — System Prompt (Condensed)}, fonttitle=\small\bfseries, left=4pt, right=4pt, top=3pt, bottom=3pt, breakable]
\small\ttfamily
You are the Strategic Director. Every K turns you set empire-level strategic direction.\par
Review the AbstractAgent's event summary and the strategic evolution chain\par
before deciding. Apply lessons from prior games; adapt to the current situation.\par
\vspace{3pt}
Strategic mode choices: EXPANSION | CONSOLIDATION | SURVIVAL | OFFENSE\par
SURVIVAL is appropriate only when a city has been lost or confirmed enemy\par
combat units are actively threatening cities. Diplomatic war status alone,\par
or sighting non-combat units, does not warrant SURVIVAL mode.\par
\vspace{3pt}
Government transition: initiate a revolution only after the target government\par
technology has been confirmed as fully researched. Do not conflate researching\par
prerequisite technologies with having unlocked the target form of government.\par
\vspace{3pt}
Output JSON: \{thoughts, current\_mode, short\_term\_goals: \{city\_building\_goal,\par
  military\_goal, tech, gov, dipl\}\}
\end{tcolorbox}

\subsection{Planner Agent ($\mathrm{LLM}_{\mathrm{plan}}$, Eqs.~\ref{eq:react}--\ref{eq:plan})}

Called every turn to translate SetGoal directives into per-city and per-unit assignments.
Has access to three on-demand query tools: \texttt{query\_city\_metrics}, \texttt{query\_army\_roster},
and \texttt{query\_tech\_status}.
The Planner decides when and whether to invoke each tool based on what information
it needs; unused tools may be skipped.

\begin{tcolorbox}[colback=green!4, colframe=green!50, title={\textbf{Planner Agent} — System Prompt (Condensed)}, fonttitle=\small\bfseries, left=4pt, right=4pt, top=3pt, bottom=3pt, breakable]
\small\ttfamily
You are the Operational Planner. Translate SetGoal objectives into concrete\par
per-city production assignments and per-unit movement directives.\par
\vspace{3pt}
Three query tools are available: query\_army\_roster returns the current unit\par
and garrison table; query\_city\_metrics returns economic breakdowns and\par
recent production history; query\_tech\_status returns\par
researched technologies and unlocked items. Invoke whichever tools are\par
needed to ground your plan in real game state before writing directives.\par
\vspace{3pt}
Two-pass planning:\par
  Pass 1 — distill empire-level needs from SetGoal into a prioritised list.\par
  Pass 2 — for each city, check its history and current state, then assign\par
  the most appropriate production item from its available build list.\par
\vspace{3pt}
Output JSON: \{thoughts, suggestion: \{city, city\_garrison\_plan,\par
  civilian\_plan, military\_plan\}\}. Max 1500 chars total.
\end{tcolorbox}

\subsection{Unit Controller ($\mathrm{LLM}_{d}$, Eq.~\ref{eq:controller})}

One instance per entity (Settler, Worker, Military unit). Receives the Planner's per-unit
directive and selects a single action from the environment's \texttt{available\_actions} list.

\begin{tcolorbox}[colback=orange!5, colframe=orange!50, title={\textbf{Unit Controller} — System + Chat Prompt (Condensed)}, fonttitle=\small\bfseries, left=4pt, right=4pt, top=3pt, bottom=3pt, breakable]
\small\ttfamily
You are a tactical EXECUTOR. Your SOLE purpose is to EXECUTE the advisor's plan.\par
NO STRATEGIC THINKING. ACTION OVER INACTION. STRICT BREVITY (reasoning $<$15 words).\par
\vspace{3pt}
For Settlers --- Speed over perfection:\par
  Step 1: Distance $\geq$4 tiles from nearest city? $\rightarrow$ Proceed to Step 2.\par
  Step 2: Tile quality (Excellent/Good/Acceptable)? $\rightarrow$ BUILD NOW.\par
  Rule: Moved 5+ turns without building? $\rightarrow$ BUILD NOW regardless.\par
\vspace{3pt}
For Military --- Fortify rules:\par
  fortify ONLY valid if unit has remaining MP AND has NOT moved this turn.\par
  If fortify NOT in available\_actions $\rightarrow$ use keep activity. NEVER retreat.\par
\vspace{3pt}
Output JSON: \{``thoughts'': \{``reasoning'': ..., ``plan'': ...\}, ``action'': ``<exact string>''\}
\end{tcolorbox}

\subsection{SceneGraph Language Rendering Example}

Below is a representative \texttt{SceneGraph.invoke()} output for a Settler at turn~42.
The graph is queried with entity ID \texttt{unit\_101} and returns a compact spatial description
that is appended to the controller's observation:

\begin{tcolorbox}[colback=gray!5, colframe=gray!45, title={\textbf{SceneGraph Output} — Settler (ID: 101) at Turn 42}, fonttitle=\small\bfseries, left=4pt, right=4pt, top=3pt, bottom=3pt]
\small\ttfamily
You are a Settler. Empire Stronghold: your nearest friendly city Rome (ID: 116)\par
medium (7 tiles) to the South. allied city Athens (ID: 120) far (12 tiles) to the\par
SouthEast. Minor Tribe Village (ID: hut\_14\_22) close (2 tiles) to the NorthEast.\par
allied Warrior (ID: 52) close (1 tile) to the East.
\end{tcolorbox}

\noindent
The \emph{Empire Stronghold} tag (global-anchor edge) prevents the Settler from drifting away
from the empire even in unexplored territory. The Minor Tribe Village entry (within radius-8 hut
edge) prompts the controller to prioritize capturing the bonus before continuing expansion.
All distance descriptions use the discretized levels (\emph{close}/\emph{medium}/\emph{far})
defined in Appendix~\ref{app:impl}.


\subsection{City Controller ($\mathrm{LLM}_{\mathrm{city}}$, Eq.~\ref{eq:controller})}

One instance per city; receives the Planner's per-city production directive and
selects from the available action list.
Implements a four-layer decision authority: (1)~obey Planner by default;
(2)~anti-switching (avoid production switch if shields $>0$ unless emergency
keyword or enemy visible); (3)~quality override (item unavailable, starvation,
or disorder); (4)~autonomous fallback in priority order.

\begin{tcolorbox}[colback=yellow!4, colframe=yellow!60!orange, title={\textbf{City Controller} --- System Prompt (Condensed)}, fonttitle=\small\bfseries, left=4pt, right=4pt, top=3pt, bottom=3pt, breakable]
\small\ttfamily
You are the City Governor. Execute the Planner's assignment while protecting\par
the city from starvation, disorder, and ungarrisoned threats.\par
\vspace{3pt}
Layer 1 --- OBEY PLANNER: Build what the Planner assigns.\par
Layer 2 --- ANTI-SWITCHING: Do NOT change production if shields > 0,\par
  UNLESS emergency (DEFEND/THREAT keyword, enemy visible, city loss imminent).\par
Layer 3 --- QUALITY OVERRIDE: Switch if item not in available\_actions,\par
  causes Food Surplus < 0, or city is in Disorder.\par
Layer 4 --- AUTONOMOUS FALLBACK: Available Actions.\par
\vspace{3pt}
Output JSON: \{``thoughts'': ..., ``action'': ``<exact string>''\}
\end{tcolorbox}

\subsection{Abstract Agent ($\mathrm{LLM}_{\mathrm{sum}}$, Eq.~\ref{eq:sum})}

Invoked every $K$ turns to compress recent SQLite action logs into a structured
semantic summary and evaluate short-term goal completion per domain.
Outputs corrective directives that ground the subsequent \texttt{SetGoalAgent} call.

\begin{tcolorbox}[colback=purple!4, colframe=purple!45, title={\textbf{Abstract Agent} --- System Prompt (Condensed)}, fonttitle=\small\bfseries, left=4pt, right=4pt, top=3pt, bottom=3pt, breakable]
\small\ttfamily
You are the Strategic Supervisor. Summarize recent events and evaluate\par
short-term goals for each domain (city, unit, tech, dipl, gov).\par
\vspace{3pt}
Priority analysis order:\par
  0. Diplomatic Events --- check first: wars declared, treaties, alliances.\par
  0.5. Threat Analysis --- lost city = CRITICAL; distinguish combat vs.\par
    non-combat units before raising military alert.\par
  1. Root Cause --- tax rate, resource allocation, unit utilization.\par
  2. Production \& Garrison Verification --- compare city status to goals.\par
\vspace{3pt}
Goal status: Completed / In Progress / Failed / Failed (Timeout).\par
Failed goal MUST trigger ``URGENT RETRY: <Goal>'' in corrective\_actions.\par
Output JSON: \{thoughts: \{summary, corrective\_actions\}, goal\_evaluations: \{...\}\}
\end{tcolorbox}

\subsection{Game Analyst ($\mathrm{LLM}_{\mathrm{post}}$, Eq.~\ref{eq:post})}

Invoked once at game end. Generates the \texttt{game\_analysis\_report.json} for the
episode by retrieving and analyzing the action logs from the SQLite database.

\begin{tcolorbox}[colback=red!3, colframe=red!40, title={\textbf{Game Analyst} --- System Prompt (Condensed)}, fonttitle=\small\bfseries, left=4pt, right=4pt, top=3pt, bottom=3pt, breakable]
\small\ttfamily
You are a Senior Forensic Game Analyst. Perform a rigorous post-mortem\par
to extract high-leverage strategic lessons via Chain-of- Success/Failure traces.\par
\vspace{3pt}
Eight-Dimension Analysis:\par
  1. Garrison \& Defense: coverage, Wall-Unit synergy, redundant units.\par
  2. Production: mix (Settler/Military/Infrastructure), analysis.\par
  3. Worker Tile Improvement: road network, irrigation timing, idle workers.\par
  4. Tax Rate: Luxury timing (wasted under Despotism), disorder response.\par
  5. Technology Path: full chronological sequence, gov-prerequisite trace.\par
  6. Government Transition: Despotism penalty quantification, transition pace.\par
  7. Over-Militarization vs. Zero-Growth Economy.\par
  8. Expansion Stagnation: pace, city spacing, missed safe-land settlement.\par
\vspace{3pt}
Recommendations must form a coherent Rapid Development Pipeline:\par
  City founding -> precise defense (Walls + military unit) -> core infrastructure.\par
Output JSON: GameAnalysisResponse \{final\_state, success,root\_causes, recommendations, ...\}
\end{tcolorbox}

\newpage

\end{document}